\RequirePackage{fix-cm}

\documentclass[smallcondensed]{svjour3}
\usepackage{natbib}
\usepackage[T1]{fontenc}
\usepackage{graphicx}
\usepackage{amsmath}
\usepackage{amssymb}
\usepackage[vlined,boxed,ruled]{algorithm2e}
\usepackage{array}
\usepackage{url}
\usepackage{float}
\usepackage{xcolor}
\usepackage{hyperref}
\usepackage{gensymb}
\usepackage{textcomp}
\usepackage{pdflscape}
\usepackage{subfigure}
\usepackage{booktabs}
\usepackage{siunitx}
\usepackage{makecell}
\usepackage{tabularx}
\usepackage{ulem}
\usepackage{booktabs}
\usepackage[toc,page]{appendix}
\usepackage{multirow}
\usepackage{multicol}

\let\cite\citep

\definecolor{Urban high}{HTML}{ff3287}
\definecolor{Urban low}{HTML}{ff00ff}
\definecolor{Industrial}{HTML}{ffcaff}
\definecolor{Parking}{HTML}{535353}
\definecolor{Road}{HTML}{000000}
\definecolor{Rapeseed}{HTML}{a800ff}
\definecolor{Wheat Barley}{HTML}{ffff00}
\definecolor{Barley}{HTML}{ffb900}
\definecolor{Peas}{HTML}{fe7141}
\definecolor{Soy}{HTML}{ff3333}
\definecolor{Sunflower}{HTML}{bd208f}
\definecolor{Corn}{HTML}{880000}
\definecolor{Corn silage}{HTML}{883f00}
\definecolor{Rice}{HTML}{ff0000}
\definecolor{Beetroot}{HTML}{82002d}
\definecolor{Potatoes}{HTML}{ffff64}
\definecolor{Grassland}{HTML}{beff13}
\definecolor{Orchards}{HTML}{0ab400}
\definecolor{Vineyards}{HTML}{0a7500}
\definecolor{Deciduous}{HTML}{1e5500}
\definecolor{Coniferous}{HTML}{003900}
\definecolor{Lawn}{HTML}{00ff00}
\definecolor{Woodlands}{HTML}{824312}
\definecolor{Minerals}{HTML}{d9d9d9}
\definecolor{Sand}{HTML}{ffffc9}
\definecolor{Snow}{HTML}{ffffff}
\definecolor{Peat}{HTML}{784e00}
\definecolor{Marshland}{HTML}{787800}
\definecolor{Intertidal zone}{HTML}{c9ffff}
\definecolor{Water}{HTML}{0000ff}

\newcommand\numberthis{\addtocounter{equation}{1}\tag{\theequation}}
\newcommand{\methodname}{Sourcerer}

\newcommand{\revision}[1]{{\color{black}#1}}

\newcommand{\repo}{\url{https://github.com/benjaminmlucas/sourcerer}}

\newcolumntype{"}{@{\hskip\tabcolsep\vrule width 1pt\hskip\tabcolsep}}

\definecolor{citecol}{rgb}{0,0,0.5}
\hypersetup{urlcolor=citecol,linkcolor=black,citecolor=citecol,colorlinks=true}

\title{A Bayesian-inspired, deep learning-based, semi-supervised domain adaptation technique for land cover mapping}
\titlerunning{\methodname}

\author{Benjamin~Lucas \and Charlotte~Pelletier \and Daniel~Schmidt \and Geoffrey~I.~Webb \and Fran\c{c}ois~Petitjean}

 \institute{Benjamin Lucas, Daniel Schmidt, Geoffrey I. Webb, and Francois Petitjean \at
                Faculty of Information Technology\\
                25 Exhibition Walk\\
                Monash University, Melbourne \\
                VIC 3800, Australia\\
                \email{\{benjamin.lucas,daniel.schmidt,geoff.webb,francois.petitjean\}@monash.edu}
            \and
            Charlotte Pelletier \at
                IRISA, UMR CNRS 6074\\
                Univ. Bretagne Sud\\
                Campus de Tohannic, BP 573\\
                56 000 Vannes, France\\
                \email{charlotte.pelletier@univ-ubs.fr}
}

\date{Received: date / Accepted: date}

\begin{document}
\sloppy

\maketitle

\begin{abstract}
Land cover maps are a vital input variable to many types of environmental research and management.
While they can be produced automatically by machine learning techniques, these techniques require substantial training data to achieve high levels of accuracy, which are not always available.
One technique researchers use when labelled training data are scarce is domain adaptation (DA)---where data from an alternate region, known as the source domain, are used to train a classifier and this model is \textit{adapted} to map the study region, or target domain.
The scenario we address in this paper is known as semi-supervised DA, where some labelled samples are available in the target domain.

In this paper we present \methodname, a Bayesian-inspired, deep learning-based, semi-supervised DA technique for producing land cover maps from \revision{Satellite Image Time Series (SITS)} data.
The technique takes a convolutional neural network trained on a source domain and then trains further on the available target domain with a novel regularizer applied to the model weights.
The regularizer adjusts the degree to which the model is modified to fit the target data, limiting the degree of change when the target data are few in number and increasing it as target data quantity increases.

Our experiments on Sentinel-2 time series images compare \methodname\ with two state-of-the-art semi-supervised domain adaptation techniques and four baseline models. We show that on two different source-target domain pairings \methodname\ outperforms all other methods for any quantity of labelled target data available.
In fact, the results on the more difficult target domain show that the starting accuracy of \methodname\ (when no labelled target data are available), 74.2\%, is greater than the next-best state-of-the-art method trained on 20,000 labelled target instances.


\end{abstract}

\section{Introduction}
\label{sec:intro}

Land cover maps enable us to observe and understand the evolution of the Earth over many spatial and temporal scales~\cite{Turner2007}, and as such, they are considered a vital component of all types of environmental research and management~\cite{Bojinkski2014,Loveland2000,Lavorel2007,Asner2005,Armsworth2006}.

\begin{figure}
\centering
\includegraphics[width=.95\linewidth]{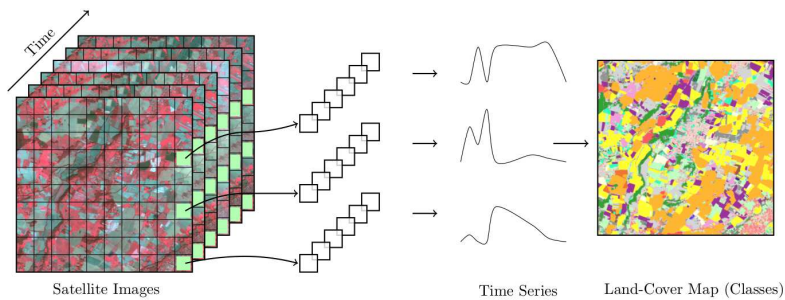}
\caption{The production of time series data from satellite images \cite{tan2017}.}
\label{fig:earth}
\end{figure}

\revision{Land cover maps can be automatically produced by applying supervised machine learning models to images acquired by satellites.
Traditionally, models were learnt from single images; however in recent times the use of temporally ordered sequences of images--- known as satellite image time series (SITS)---has become the new standard~\cite{Inglada2017}.
Figure~\ref{fig:earth} depicts the production of time series from a pixel of Earth imaged by satellite.
Each pixel is recorded as a decimal value on multiple spectral bands, and given that this occurs at repeated time intervals, the result is a multivariate time series for each pixel that we can use for classification.}

Maps produced using SITS have been found to be significantly more accurate, as these data enable classification of some land cover types that single images do not~\cite{defourny2019,vuolo2018}; for instance, soy and corn are both winter crops and will appear similar in a single image.
In contrast, their different growth rates will be clearly evident using SITS data.

\begin{figure}
\centering
\includegraphics[width=.9\linewidth]{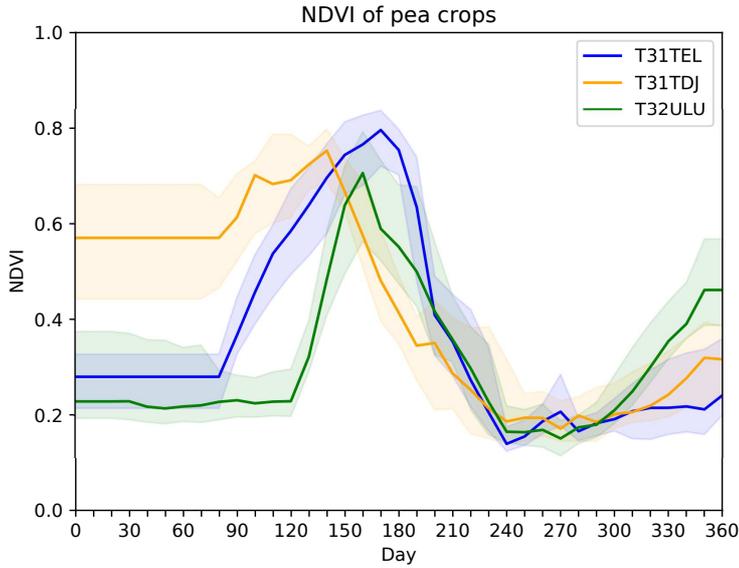}
\caption{Mean NDVI and quartiles of pea crops in 2016 for three Sentinel-2 satellite tiles located in France (see Figure~\ref{fig:map} for the exact locations of the tiles).}
\label{fig:NDVI_eg}
\end{figure}

The current state-of-the-art methods for producing land cover maps from SITS are deep learning and random forests~\cite{Wulder2018,azzari2017}.
However, the accuracy of both of these is highly dependent upon the availability of a large quantity of labelled data. The need for large quantities of labelled training data presents a major problem in land cover mapping for three reasons:
\begin{enumerate}
    \item Labelled data are both expensive and time-consuming to acquire at the resolution of the latest Earth observation satellites (10 meters in the case of Sentinel-2 as used in this paper).
    \item The data are often specific to their location. For example, Figure~\ref{fig:NDVI_eg} shows the mean Normalized Difference Vegetation Index of pea crops growing in 3 different regions of France. It is clear that even within one country, the same crop can take on three distinctly different profiles, meaning we cannot simply \textit{borrow} data from a nearby region when training a model. \revision{Agricultural practices, water, soil, weather and many other factors can also contribute to variation between spectral profiles of crops}.
    \item Land cover changes over time, and thus we cannot reliably use old labelled data to train a new model as it may no longer be accurate.
\end{enumerate}
Consequently, labelled data which is both recent and sourced from the study area are at best scarce, and frequently non-existent, making utilizing the state of the art to create land cover maps extremely challenging.

Researchers have proposed three main approaches to tackle this problem: (1)~using out-of-date reference data \cite{Tardy2017}; (2)~active learning strategies \cite{persello2012active,matasci2012active}; and (3)~domain adaptation (DA).

The first approach best suits the scenario in which accurate historical data exist (which is often not the case for the same reasons outlined above) and state-of-the-art algorithms can be used to identify which of the historical data are now outdated~\cite{frenay2014,damodaran2020,pelletier2017filtering,bailly2018}.
The second approach best suits the scenario where sufficient computational resources are available and where further data collection is feasible (i.e., timely, affordable)~\cite{Tuia2011}, and therefore  lends itself favorably to smaller scale applications.
The third approach, DA, is best suited to a situation where ample labelled data from a different location are readily available to the practitioner and can be used to train a classifier.
Some DA approaches are specifically aimed at the scenario where some additional labelled data are available from the study area, which is the one we present in this paper.

\revision{
In DA, a labelled dataset---the \textit{source domain}---is utilized for the purpose of classifying instances from a dataset where labels are scarce or unavailable---the \textit{target domain}.
The main challenge of DA is the variation in feature distributions between the source and target domains.}
Generally speaking, DA methods can approach this in two ways: (a)~by \textit{adapting} the source domain data to appear more statistically similar to the target domain; or (b)~by learning a classifier on the source domain data and then \textit{adapting} it to classify the target domain.

In this paper, we address the particular scenario of semi-supervised DA, in which a relatively small amount of labelled data is available from the target domain.
Our method, \methodname, is a deep learning-based method for semi-supervised DA based on a Bayesian-inspired, novel regularizer for the weights of a convolutional neural network (CNN).
We demonstrate \methodname\ on Sentinel-2 image time series data and show that on a 30-class land cover classification problem it outperforms the current state-of-the-art methods in semi-supervised DA, regardless of the given quantity of labelled target data available.
In particular, our contributions can be summarized as follows:
\begin{enumerate}
    \item Proposing \methodname: a novel method for semi-supervised DA on SITS data;
    \item Achieving state-of-the-art performance on two separate source-target pairings on Sentinel-2 data;
    \item Providing a semi-supervised DA method emphasizing a user-friendly implementation, as it can be applied to a pre-trained deep learning model and does not require the user to possess the source domain training data;
    \item Providing an open source implementation of \methodname\ for reproducibility and wider implementation.
\end{enumerate}

The remainder of the paper is organized as follows: Section~\ref{sec:da} discusses DA, the current state of the art, and presents the existing work using DA for remote sensing; Section~\ref{sec:ourmethod} presents \methodname: our Bayesian-inspired, deep learning-based method for semi-supervised DA; Section~\ref{sec:data} details the data used in the experiments presented in Section~\ref{sec:exp}; finally, we draw conclusions and suggest future directions in Section~\ref{sec:con}.

\section{Domain Adaptation}
\label{sec:da}
DA belongs to a family of machine learning techniques that deals with data distributions that are not stationary over time or space~\cite{Tuia2016}.
It utilizes labelled data from a \textit{source domain}, in which labels are widely available, for the purpose of classifying the area of interest in which labels are scarce (or unavailable), the \textit{target domain}.
Implicit in this are the assumptions that the source joint distribution $p_{source}(X,Y)$ is sufficiently different to the target joint distribution $p_{target}(X,Y)$ for it to be sub-optimal to use a model trained on $p_{source}$, but nonetheless sufficiently similar to be useful for the learning task, where $X$ is the input (observations) and $Y$ the output (land cover labels).

When using SITS for land cover mapping, DA can be applied in two ways: temporally or spatially.
The first situation primarily arises when a map is in need of updating but reference data from the present time is unavailable.
In this case, a map of the study area from a previous year (or years) can be used as the source domain and adapted to map the present day land cover (the target domain)~\cite{Tardy2017,Demir2013,tardy2019otlandcover}.

The second setting, and the one we will explore in this paper, occurs when data from one geographical region is used as the source domain and DA is used to map a different geographical region (the target domain).
\revision{In this paper, we have chosen to demonstrate our method using this setting because of the lack of existing research in this area using SITS; however our method is equally applicable to temporal DA.}

There are two general scenarios that are presented in DA research---\textit{unsupervised} DA and \textit{semi-supervised} DA~\cite{kouw2019dareview}---which differ in whether labelled target data is available.
In unsupervised DA, no labelled data are available in the target domain and the methods acquire information only from the structure of the unlabelled data.
In semi-supervised DA, some labelled samples are available. However, there are usually insufficient samples to train an accurate classifier, so the labelled target data works to complement the source data in training a classifier.\footnote{We note that this differs from the definition given in the most cited survey of DA in remote sensing~\cite{Tuia2016} but is consistent with the definition used in the overwhelming majority of DA research, particularly in the field of computer vision~\cite{patel2015}.}
In accordance with the definition of DA, it assumed that sufficient labelled source data is available in both scenarios.

While the vast majority of DA research focuses on unsupervised methods, we have chosen to present a semi-supervised DA method as we believe that this is a more practical scenario in remote sensing/land cover mapping---where funding is available to obtain \textit{some} labelled data.
In this case, a state-of-the-art semi-supervised DA technique would help practitioners produce high-accuracy land cover maps without having to perform additional large scale data collection.

In the following section we provide a brief overview of the state of the art in both unsupervised and semi-supervised DA.

\subsection{Unsupervised Domain Adaptation}
Due to the large quantity of research in unsupervised DA, we emphasize the current state of the art; for a more comprehensive review of the field we direct the reader to~\citet{kouw2019review}. Unsupervised DA occurs when labelled data is available in the source domain, while the target domain has only unlabelled samples available.
Early techniques addressing this problem attempt to align the source and target data spaces, or projections thereof, to one another \cite{huang2007iwkmm,kouw2016flda}.
These methods often also include use of dimension reduction techniques, such as principal component analysis or transfer component analysis, based on the assumption that the reduced spaces will be more similar to one another \cite{pan2011tca,fernando2013sa,gong2012geodesic}.
These ideas have been further extended in~\citet{long2015learning} by the addition of deep learning and the maximum mean discrepancy criteria to find features that are transferable between domains.

More recently, DA research has had a marked shift towards deep learning methods.
The primary difference is that traditionally, the adaptation method and the classifier used to be orthogonal to one another,
deep learning-based methods perform the adaptation and the training of the classifier in one step (often simultaneously). Deep learning methods have been applied in various ways, including: sharing model weights~\cite{sun2016deepcoral}; adversarial loss functions~\cite{ganin2016dann}; generative adversarial networks~\cite{tzeng2017adda}; and iteratively learning the target-domain decision boundary~\cite{shu2018dirt}.

Another major area of recent research in unsupervised DA is optimal transport (OT), which seeks to find the minimum optimal transformation $(T)$ between the source and target distributions by attributing a cost to the transformation of each instance in the dataset~\cite{courty2016ot,damodaran2018deepjdot}.
OT has been used in land cover mapping to produce maps with no present day reference data, where maps from previous years are used as the source domain and the present day land cover used as the target.
~\citet{tardy2019otlandcover} found that a 17-class problem was too difficult for most variants of OT, with the best producing a map with only 70 percent accuracy.
It has also been shown that OT can be used in a multimodal context for land cover mapping---where data from one device acts as the source domain and another the target~\cite{courty2016otremotesensing}.

\subsection{Semi-supervised Domain Adaptation}
Semi-supervised DA occurs when labelled data is available in both the source and target domains, but the quantity available in the target domain is insufficient to train an accurate model.

This scenario has great applicability to land cover mapping as resources are rarely available for a large scale data collection campaign, and therefore a successful semi-supervised DA method will allow for the production of large-scale maps at a small fraction of the cost.

For example, \citet{Inglada2017} required approximately 35 million training instances to create a land cover map of France, a quantity that is unfeasible to obtain in many nations, particularly those that are resource-poor.

While less studied, semi-supervised DA research has followed a similar trajectory to unsupervised DA research over the last decade.
In fact, a number of unsupervised methods have also been applied to the semi-supervised setting with slight modifications to utilize the labelled target data.

Most early methods worked by mapping the source and target domain data to a new feature space, ensuring that instances from the same class map to a similar area of the space (regardless their originating domain)~\cite{gong2012geodesic,wang2011ma}. 
In general, these methods do not handle non-linear deformations or high-dimensional data problems particularly well and are therefore of less relevance to the field of remote sensing~\cite{tuia2016kema}.

Kernel manifold alignment (KEMA)~\cite{tuia2016kema} was developed to combat the issue of dealing with high-dimensional data, by creating the data transform based on only a few labelled samples from each domain.
However when KEMA was used in a land cover mapping problem by~\citet{bailly2017} the results were unsatisfactory, yielding only 70 percent accuracy on a 7-class classification problem.

Recently, deep learning has resulted in marked advances in semi-supervised DA.
Domain-adversarial neural networks (DANN)~\cite{ganin2016dann} can be used as either a semi-supervised or unsupervised method, as required. This method aims to learn class labels that are domain-independent.
To achieve this, a CNN is trained with a loss function comprised of two components---a class-specific component and a domain-specific component.
This approach seeks to simultaneously minimize the loss of predicting class labels while maximizing the loss of predicting whether the instance came from the source or target domain.
Consequently, the model learns to accurately classify classes while having increasing difficulty distinguishing between domains.
\revision{DANN has been successfully applied to land cover mapping in the unsupervised DA case~\cite{bejiga2019} but its method of aligning domains has been shown to be less successful in the semi-supervised case~\cite{saito2019semi}.}

The other method representing the current state of the art in semi-supervised DA is minimax entropy (MME)~\cite{saito2019semi}.
This method learns a prototype (a representative datapoint) for each class in the labelled data and then minimizes the distance between these prototypes and the unlabelled data, thus learning discriminating features.
As the labelled data is dominated by instances from the source domain, the method uses a novel adversarial method to shift the class prototypes towards the target domain data.
\revision{This shift means that MME performs best in cases where the classes are represented approximately evenly in the labelled target data, and particularly in the case when they can be chosen selectively.}

\revision{Neither state-of-the-art method has seen a wide uptake in land cover mapping.
This may be a result of a lack of performance in this specific application, or it may also be due to practical reasons such as data storage or computational cost.
Both DANN and MME perform semi-supervised DA by first pooling the labelled source and target data, thus in all cases the labelled source data (often 10M+ instances) must be available.}
This is a major point of difference with the approach we propose in this paper.

\section{\methodname}
\label{sec:ourmethod}
The method we present in this paper, \methodname, is a novel, Bayesian-inspired, deep learning-based method for semi-supervised DA for SITS.
It uses a CNN model trained on the source domain as a starting point and improves upon it by further training it on the available labelled target data.
A critical and distinguishing feature of our approach is a novel regularizer (\text{SourceRegLoss}) that is used while training on the target data. This tunes the amount of \textit{trust} placed on the updates.
That is, as the quantity of labelled target data increases, the model places gradually more trust on what is learnt from this data (and consequently, relies less upon the weights learnt from the source data).

\methodname\ not only delivers excellent performance, but is also widely applicable as it does not require access to the source data, but instead works on a model previously trained on that data.
In our experiments, we demonstrate the flexibility of our approach by training a model on a source domain once, and then utilizing this pre-trained model and \methodname\ to classify two different target domains.

\subsection{TempCNN}
\label{subsec:tempcnn}
The CNN model utilized by \methodname\ is TempCNN~\cite{Pelletier2019}, which has been shown to be a highly accurate model for pixel-based analysis of SITS data.
It has been demonstrated to significantly outperform other types of deep learning models, including recurrent neural networks, at large geographical scale.

\begin{figure}[!htb]
\centering
\includegraphics[width=.95\linewidth]{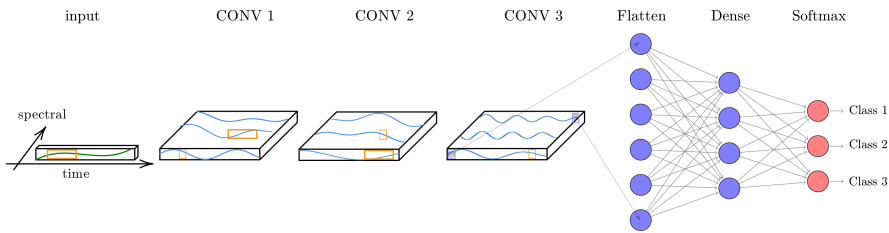}
\caption{The TempCNN model architecture, as presented in~\citet{Pelletier2019}.}
\label{fig:tempcnn}
\end{figure}

The model comprises 3 convolutional blocks, followed by 1 fully-connected block and a softmax layer~(see Figure~\ref{fig:tempcnn}).
The convolutional block consists of 64 convolutional filters of length 5, followed by a batch normalization layer, a dropout layer with a rate of 0.5, ending with a ReLU activation function.
The convolutions are 1-dimensional and are performed along the temporal axis only.
The fully-connected layer has 256 neurons, followed by the same batch normalization layer, dropout and ReLU function.
The final layer in our case is a softmax with 30 units representing the 30 land cover classes of our classification problem (see Section~\ref{subsec:refdata}).

\subsection{Source-regularized Loss Function}
\label{subsection:sourceloss}

To utilize \methodname, one must first either train a model using the labelled source data with a standard loss function (for example, categorical cross-entropy loss for a classification problem), or obtain a pre-trained model.
Let $\hat{\theta}_s$ denote the estimates of the parameters of the source model.
Then, using these estimates as a reference point, the new target model is trained on the labelled target data using the following source-regularized loss function:

\begin{equation} 
\label{sourceregloss}
{\rm SourceRegLoss}(X_t,y_t,f_\theta,\hat{\theta}_s,\lambda) = L\big(f_\theta(X_t),y_t\big) + \lambda ||\theta - \hat{\theta}_s||^2
\end{equation}
where:\\
\hspace*{5mm} $L$ is the average loss (calculated per sample);\\
\hspace*{5mm} $f_\theta$ is the current model with parameters $\theta$;\\
\hspace*{5mm} $X_t, y_t$ are the target data and labels, respectively;\\
\hspace*{5mm} $\hat{\theta}_s$ are the estimated parameters of a model trained on the source data; and \\
\hspace*{5mm} $\lambda$ is the regularization hyperparameter.

\noindent During training, the proposed regularizer acts to \textit{shrink} the values of the estimated parameters towards those that were learned on the source data. This is done by adding the squared difference between the  parameters of the target model, $\theta$, and the estimated parameters of the source model, $\hat{\theta}_s$, to the loss function, penalizing parameter estimates that deviate substantially from the source model. This approach is motivated by the more general ideas of Bayesian inference. In Bayesian inference one formally specifies a prior guess at the likely population values of a model through the mechanism of a prior distribution. The resulting posterior distribution combines the information contained in the sample with the information in the prior. In our case, the use of the source model parameter estimates $\hat{\theta}_s$ as a reference point mimics the use of a prior distribution. The correspondence is even closer than this, due to the relationship between squared-penalties and normal distributions (see Section \ref{subsec:bayes_connections} for further discussion). A similar approach has been previously used in transfer learning~\cite{dalessandro2014,aljundi2018memory}, but to the best of our knowledge this is the first time it has been adapted for time series classification, the field of Earth observation and to CNNs in general.

The hyperparameter $\lambda$ controls the degree to which deviations of the target model from the source model are penalized. In standard Bayesian inference, the information contained in the prior distribution is outweighed by the information contained in the sample as the sample size grows, so that the effects of regularization are greater for small amounts of target data and correspondingly reduced as the amount of target data increases. We discuss in Section \ref{subsec:lambda} a simple technique for choosing $\lambda$ that mimics this behavior. The regularization is applied to all learnable parameters of the model; for the TempCNN this includes the weights and biases of the convolutional layers, the fully-connected layers, and the batch normalization layers.

We note that we found optimum results by freezing the running mean and running variance parameters of the batch normalization layer after training on the source data.
This is because the available target data have low variability as they are from a limited number of polygons and therefore the batch mean and batch variance of these data are not representative of the data as a whole. That is, the batch means are skewed towards the classes present and the batch variance will be lower given the limited number of classes present in the target data.

It is also important to emphasize that our proposed loss function adds no additional computational cost to the training of the target model.

\subsection{Determining the Regularization Hyperparameter}
\label{subsec:lambda}

The amount of regularization applied by \methodname\ is determined entirely by the choice of $\lambda$. In this section we propose a heuristic choice that automatically balances the amount of regularization against the amount of available labelled target data. When the quantity of labelled target data is small, we would like the procedure to use a large value for $\lambda$, making the values of the weights tend toward the parameters of the source model. To see that such a schedule is sound, we note that as the amount of target data, $n_t$, grows the average loss is of order $O(1)$, i.e., it does not grow in magnitude with $n_t$. 
To ensure that for large amounts of target data the regularization has little effect we require that $\lambda(n_t) = o(1)$, i.e., tends to zero as $n_t \to \infty$. This is a necessary condition for our learning procedure to be statistically consistent.

To achieve this desired behavior, we propose a simple heuristic schedule for $\lambda$. We fix the value of $\lambda$ at the two extreme points: (i) when we have a minimum quantity of target data ($t_{min}$, $\lambda_{t_{min}}$) and (ii) when we have some large amount of target data ($t_{max}$, $\lambda_{t_{max}}$). We then fit a concave-up power curve between these points. The usual form of a power curve is:
\begin{equation} \label{power_curve}
\lambda = A \, {n_{t}}^{-k}
\end{equation}
where:\\
\hspace*{5mm} $\lambda$ is the regularization hyperparameter; \\
\hspace*{5mm} $n_{t}$ is the quantity of labelled target data available; and, \\
\hspace*{5mm} $A$, $k$ are constants. \\

Using the properties of a power curve, this formula can also be represented as a linear equation on a log-log scale.
Therefore, to find the schedule for $\lambda$ we find  the line that passes through the log transform of our two points: (log($t_{min}$), log($\lambda_{t_{min}}$)) and (log($t_{max}$), log($\lambda_{t_{max}}$)), respectively.
The slope of the resulting line is:
\begin{equation} \label{slope}
k = \frac{\log \lambda_{t_{max}} - \log \lambda_{t_{min}} }{\log t_{max}-\log t_{min}}.
\end{equation}
Using this slope and the point ($\log t_{min}$, $\log \lambda_{t_{min}}$), we can define the equation of the line as:
\[
    \log \lambda - \log \lambda_{t_{min}} = k \left( \log n_{t} - \log t_{min} \right)
\]
and solve for $\lambda$, yielding
\begin{equation}
    \lambda = \left(\frac{\lambda_{t_{min}}}{{t_{min}}^k}\right){n_{t}}^k \numberthis \label{line}.
\end{equation}
We now describe some simple and reasonable heuristic choices for some of the free variables. In the (unlikely) case in which only one labelled target instance is available, a very large value for $\lambda$ will ensure that the model uses the source parameters. By similar reasoning, when a significant amount of labelled target data is available, a suitably small value of $\lambda$ will allow the model to learn from the target data and largely ignore the source model. Following this argument, we set $t_{min} = 1$, $\lambda_{t_{min}} = 10^{10}$, and $\lambda_{t_{max}} = 10^{-10}$ and Equation~\ref{line} reduces to:

\begin{equation}
\lambda = 10^{10} {n_{t}}^k \numberthis \label{final}
\end{equation}
where $k$ is now
\begin{equation}
k = -\frac{20 \log 10 }{\log t_{max}}. \numberthis \label{final_slope}
\end{equation}
This leaves $t_{max}$ as the only free, user-specified hyperparameter of the procedure. We note that as long as $t_{max} > 1$ (a reasonable choice), then $k<0$ and the schedule (\ref{final}) satisfies the condition $\lambda = o(1)$, as prescribed above. We have performed a sensitivity analysis of this parameter in Section~\ref{subsubsec:sensitivity}.

\subsection{Connection to Bayesian Inference}
\label{subsec:bayes_connections}

We now examine the close connection between \methodname\ and Bayesian inference. This has been previously noted, but we now make the connection more explicit. First we briefly review Bayesian statistics. In the Bayesian approach we have a probabilistic model of data, $p(y | \theta)$, with unknown parameters $\theta$ that we would like to fit to some observed dataset. We further must propose a probability distribution $\pi(\theta)$ that describes our belief about which values of $\theta$ are likely to be the (unknown) population value, before seeing the data (i.e., \textit{a priori}). This is called a prior distribution. Bayesian inference proceeds by forming a posterior distribution using Bayes' rule:
\[
    p(\theta | y) = \frac{p(y | \theta) \pi(\theta)}{p(y)},
\]
where $p(y)$ denotes the marginal distribution of the data. The posterior distribution describes the likelihood of certain values of $\theta$ being the true (unknown) population value of $\theta$, after observing data $y$, and is used as a basis for statistical inference. In practice, computing the normalizing term $p(y)$ is usually infeasible, particularly for complex models such as neural networks, and instead of using the complete posterior it is common practice to estimate $\theta$ by maximizing the unnormalized posterior
\[
    \hat{\theta} = \arg \max_{\theta} \left\{ p(y | \theta)\pi(\theta) \right\}.
\]
A particular strength of the Bayesian framework is that it allows us to formally encode our prior beliefs, or previous information, into the learning process. 

We can connect \methodname, and the source-regularized loss (Eq.~\ref{sourceregloss}) on which it is based, to Bayesian inference by noting several equivalencies. First, we note that maximizing the posterior is equivalent to minimizing the negative logarithm of the posterior. The choice of cross-entropy loss for categorical regression is equivalent to choosing our data model $p(y | \theta)$ to be an appropriate neural network with a multinomial logistic regression output layer, and our choice of $\ell_2$ regularization is equivalent to assuming a normal prior distribution for the parameters of the form
\[
    \theta_j \sim N \left( [ \hat{\theta}_s ]_j, \frac{2}{\lambda} \right),
\]
that is, assuming that each of the model parameters is \textit{a priori} normally distributed with a mean equal to the estimated value of corresponding parameter in the source model, and a variance inversely proportional to $\lambda$. In this way we can interpret $[ \hat{\theta}_s ]_j$ as setting our ``best guess'' for the value of our parameter, and $\lambda$ as determining how much weight we place on our prior beliefs. Large values of $\lambda$ lead to small prior variance, and a concentration of probability around our prior guess $[\hat{\theta}_s]_j$, and small values spread probability more diffusely, placing less importance on our prior guess.

We note that this idea of using a prior guess and regularizing a loss for estimation of (high dimensional) parameter vectors is itself certainly not new. In fact, the concept dates back as early as the seminal work of \citet{james1961}, a ground-breaking piece of work in which the authors propose the first formal shrinkage estimator. The James-Stein procedure was designed to estimate the mean of a multivariate normal, and was shown to uniformly improve on regular least-squares (i.e., equivalent in our setting to using the target data only) by shrinking the estimates towards a reference point (equivalent to the existence of a source model). This is essentially the same idea that underlies our proposal. 

The Bayesian connection of our method also offers the possibility for further improvements to \methodname.
\citet{james1961} show that the optimal choice of $\lambda$ is inversely proportional to an unbiased estimate of the Kullback--Leibler divergence between the target only model and the reference (source) model; that is, the more the target only model differs from the reference model, the less weight should be placed on the reference.
Though accurate estimation of Kullback--Leibler divergences between neural networks is difficult, a similar idea could potentially be adapted for use in \methodname\ to refine the selection of $\lambda$.

Another way of choosing $\lambda$, would be in a more formal, and data-driven manner, with a prior distribution placed on $\lambda$, and it integrated directly into the posterior distribution.
In this manner an appropriate value for $\lambda$ could be estimated directly from the target data by a straightforward integration into a posterior sampling scheme or a variational Bayes approach, both of which are gaining popularity in the neural network community.

\section{Data}
\label{sec:data}
All experiments were performed using the SITS data acquired by the Sentinel-2A satellite, starting on 1 January 2016 and running through to 26 December 2016 (its twin satellite Sentinel-2B was launched in March 2017).
Table~\ref{tab:image_dates} shows the dates of the images for each satellite tile used in our experiments (tiles discussed further in Section~\ref{subsec:tiles}).

\begin{table}[!htbp]
    \centering
    \caption{Original image dates for each tile used in the experiments and the interpolated dates after pre-processing (all from 2016)}
    \label{tab:image_dates}
    \begin{tabular}{c|c|c|c}
        \toprule
        \textbf{T31TEL} & \textbf{T31TDJ} & \textbf{T32ULU} & \textbf{Interpolated Dates} \\
        \midrule
        12-MAR & 12-JAN & 26-JAN & 01-JAN \\
        22-MAR & 12-MAR & 05-FEB & 11-JAN \\
        08-APR & 22-MAR & 09-MAR & 21-JAN \\
        28-APR & 29-MAR & 26-MAR & 31-JAN \\
        08-MAY & 08-APR & 29-MAR & 10-FEB \\
        18-MAY & 08-APR & 08-APR & 20-FEB \\
        21-MAY & 11-APR & 28-APR & 01-MAR \\
        28-MAY & 18-APR & 05-MAY & 11-MAR \\
        07-JUN & 28-APR & 08-MAY & 21-MAR \\
        20-JUN & 01-MAY & 25-MAY & 31-MAR \\
        27-JUN & 18-MAY & 28-MAY & 10-APR \\
        30-JUN & 21-MAY & 07-JUN & 20-APR \\
        07-JUL & 28-MAY & 24-JUN & 30-APR \\
        10-JUL & 07-JUN & 24-JUN & 10-MAY \\
        17-JUL & 10-JUN & 07-JUL & 20-MAY \\
        20-JUL & 20-JUN & 17-JUL & 30-MAY \\
        30-JUL & 27-JUN & 27-JUL & 09-JUN \\
        06-AUG & 07-JUL & 13-AUG & 19-JUN \\
        16-AUG & 10-JUL & 16-AUG & 29-JUN \\
        19-AUG & 17-JUL & 23-AUG & 09-JUL \\
        26-AUG & 20-JUL & 26-AUG & 19-JUL \\
        29-AUG & 27-JUL & 02-SEP & 29-JUL \\
        05-SEP & 30-JUL & 12-SEP & 08-AUG \\
        08-SEP & 06-AUG & 22-SEP & 18-AUG \\
        25-SEP & 16-AUG & 25-SEP & 28-AUG \\
        28-SEP & 19-AUG & 02-OCT & 07-SEP \\
        05-OCT & 26-AUG & 05-OCT & 17-SEP \\
        15-OCT & 29-AUG & 12-OCT & 27-SEP \\
        18-OCT & 05-SEP & 22-OCT & 07-OCT \\
        18-OCT & 15-SEP & 22-OCT & 17-OCT \\
        18-OCT & 28-SEP & 01-NOV & 27-OCT \\
        18-OCT & 08-OCT & 01-DEC & 06-NOV \\
        07-NOV & 15-OCT & 04-DEC & 16-NOV \\
        17-NOV & 18-OCT & 11-DEC & 26-NOV \\
        27-NOV & 18-OCT & 14-DEC & 06-DEC \\
        04-DEC & 04-NOV & 21-DEC & 16-DEC \\
        07-DEC & 14-NOV & 31-DEC & 26-DEC \\
        14-DEC & 17-NOV & & \\
        17-DEC & 27-NOV & & \\
        24-DEC & 07-DEC & & \\
        27-DEC & 14-DEC & & \\
        & 17-DEC & & \\
        & 27-DEC & & \\
        \bottomrule
    \end{tabular}
\end{table}

\subsection{Preprocessing}
All Sentinel-2A data have been collected and prepared by our colleagues from the CESBIO lab using \texttt{iota2} software~\cite{iota2}. The key steps in this process are outlined below:
\begin{itemize}
    \item Atmospheric, adjacency and slope effects are corrected for using the MAJA processing chain \cite{Hagolle2015}.
    The output of this are top-of-canopy images with associated clouds masks. We note that only the images with a cloud-cover of less than 80~\% are processed by MAJA.
    \revision{\item Each image is comprised of 13 spectral bands---four of which are recorded at a spatial resolution of 10 metres; six that are recorded at a resolution of 20 metres, which are then reinterpolated at 10 metres; and three that are recorded at a resolution of 60 metres, which are discarded at this stage as they are only used in atmospheric correction and cloud detection.}
    \item The images are gapfilled using a linear temporal interpolation with a time gap of 10 days, resulting in 37 dates for each pixel \cite{Inglada2017,defourny2019}.
    Ten days is a natural choice for the time gap as it represents the revisit frequency of one Sentinel 2 satellite.
    However, the orbit of the satellite results in some \textit{overlapping} between areas and therefore some pixels are imaged more frequently than others.
    Thus, gapfilling is a vital processing step to ensure that each pixel has the same number of timestamps.
    It also allows for the correction of images that are compromised by cloud-cover.
    Table~\ref{tab:image_dates} shows the dates of the original images for each satellite tile and the interpolated dates.
\end{itemize}

After the MAJA processing, the resulting instances (pixels) are each represented by a multivariate time series with 10 variables (one for each spectral band) of length 37.
The data has been normalized per spectral band using values from the source domain data.
Following~\citet{Pelletier2019}, a variation on min-max normalization has been used, replacing the absolute minimum and maximum values with the 2nd and 98th percentile values, respectively. The percentiles used are estimated using  all of the values of the series at each individual timestep. This normalization differs from the usual method for time series classification~\cite{bagnall2017} but deliberately avoids two potential pitfalls in using standard methods.
First, it retains the relative scale of the spectral bands as this is important to SITS data (for instance, in the calculation of normalized difference vegetation index). Second, if the data were normalized per image, the ability to track changes over time would be lost. The normalization method used preserves both the capacity to combine band values and to track changes through time.

\revision{It should be noted that all of the models in this study require all data to be of the same spatial resolution and be of the same length (number of time steps).
This is a current limitation of using deep learning models on SITS data in general, and not otherwise related to \methodname.}

\subsection{Reference Data}
\label{subsec:refdata}
The reference data are the same as those used previously to produce a land cover map of France in 2016 with the methodology presented in \citet{Inglada2017}. The reference data originate from four sources:
\begin{enumerate}
    \item The Agricultural Land Parcel Information System (2016) (\textit{Registre Parcellaire Graphique}): a compilation of data gathered from farmers' declarations of agricultural land~\cite{Farmers}.

    \item Urban Atlas (2012): a land cover dataset gathered by the European Environment Agency (EEA) detailing the land cover of cities in continental Europe at a very high resolution (2.5 metres) using 27 urban classes~\cite{urbanatlas}.
    
    \item The CORINE Land Cover Inventory (CLC 2012): an inventory of land cover information gathered by the EEA using 44 land cover classes at a spatial resolution of 250 metres~\cite{CORINE}.
    
    \item French National Geographic Institute `BD-Topo': a national topographical map of produced by the government of France~\cite{bdtopo}.    
\end{enumerate}

Information from these sources has been amalgamated to create a dataset using a nomenclature of 30 land cover classes:
\begin{itemize}
    \item Five urban classes: High-density Urban, Low-Density Urban, Industrial, Parking, Roads;
    \item Fourteen vegetation classes: Rapeseed, Winter Wheat and Barley, Spring Barley, Pea, Soy, Sunflower, Corn, Corn silage, Rice, Beetroot, Potatoes, Grassland, Orchards, Vineyards;
    \item Seven natural and semi-natural classes: Deciduous Forest, Coniferous Forest, Lawn, Woodlands, Surface Minerals, Beaches and Dunes, Glaciers; and,
    \item Four \textit{other} classes: Peat, Marshland, Inter-tidal Land, Water.
\end{itemize}

\subsection{Source and Target Tiles} \label{subsec:tiles}
\begin{figure}[!htb]
\centering
\includegraphics[width=.8\linewidth]{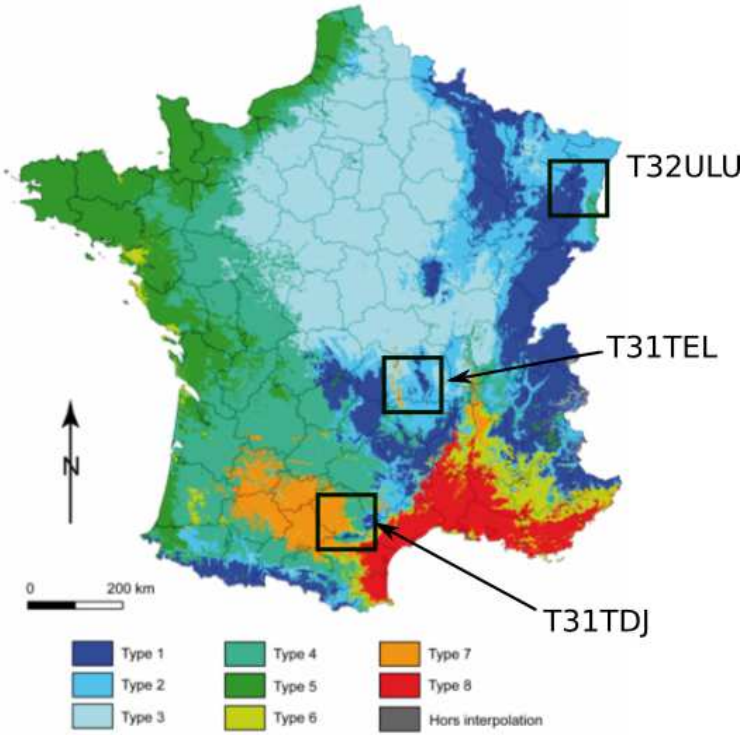}
\caption{Climate map of France from~\cite{joly2010} with our three study regions identified.}
\label{fig:map}
\end{figure}

The experiments were conducted on three study areas: we used one as a source domain and two as target domains, with each area representing a Sentinel-2 tile (110km $\times$ 110km).
The experiment regions are all located in France (see map in Figure~\ref{fig:map}) as there is full reference data available as outlined in Section~\ref{subsec:refdata}.
The source domain (tile T31TEL) was chosen at random amongst the available Sentinel-2 tiles and is located within a highland region known as Massif Central~(45.1\degree N, 2.6\degree E).
The two target tiles were chosen specifically to observe the variation in results between a target region with a similar climatic profile (T32ULU) and a target region with a very different climatic profile (T31TDJ).
Target domain T31TDJ is located near the city of Toulouse in south-west France~(43.6\degree N, 1.4\degree E), and T32ULU is located in the north-eastern region of France called Grand Est, which includes the city of Strasbourg~(48.4\degree N, 7.5\degree E).

\begin{figure}[!htb]
\centering
\includegraphics[width=1\linewidth]{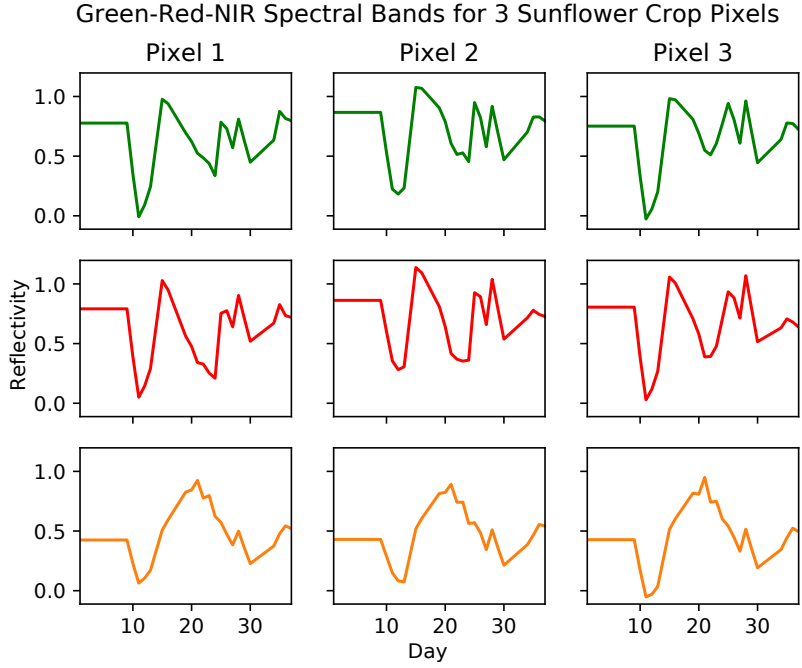}
\caption{The green, red and near infrared reflectance time series for 3 different sunflower pixels located in the same polygon.}
\label{fig:polygon_example}
\end{figure}

Our colleagues at CESBIO who provided us with the preprocessed data also provided us with predefined train and test sets per tile. We have chosen not to modify this split as it has been performed such that instances that belong to the same polygon are in the same set---ensuring independence between training and testing sets (as per~\citet{roberts2017}).
In our data, a polygon represents a contiguous area of one land cover class (a corn crop, a river, an industrial estate, etc.) and consequently instances from the same polygon have near-identical profiles.
For example, Figure~\ref{fig:polygon_example} depicts three of the spectral bands from three different pixels of sunflower from within the same polygon.
The similarity between these instances demonstrates that if the data were split at random, rather than blocked by polygon, and these instances were distributed to both the train and test sets, the problem of classifying them would be trivial.

Table~\ref{tab:no_instances} displays the total number of instances per domain and per set.
We note that while this shows \textit{all} of the target training data, we conduct our experiments under the condition that only a predetermined quantity is available (per experiment), and we study the evolution of test accuracy for increasing quantities of target training data.

\begin{table}[!htbp]
    \centering
    \caption{Total number of train and test instances (pixels) available for each domain.}
    \label{tab:no_instances}
    \begin{tabular}{rccc}
        \hline
         & \textbf{Source} & \textbf{Target 1} & \textbf{Target 2}\\
         & T31TEL & T31TDJ & T32ULU \\
         \hline
        \textbf{Train} & 12,647,452 & 8,758,196 & 15,122,125 \\
        \textbf{Test} & - & 3,371,843 & 5,599,461 \\
        \hline
    \end{tabular}
\end{table}

A comparison of the land cover classes of the training data for the regions is displayed in Table~\ref{tab:class_dist}.

\begin{table}[!htbp]
    \centering
    \caption{Distribution of land cover classes across each domain.}
    \label{tab:class_dist}
    \begin{tabular}{c|c|r|r|r}
        \toprule
        Label & Description & \makecell{\textbf{Source} \\ T31TEL} & \makecell{\textbf{Target 1} \\ T31TDJ} & \makecell{\textbf{Target 2} \\ T32ULU} \\
        \midrule
        1 & Urban (high density) & 16709 (0.13\%) & 18242 (0.14\%) & 9871 (0.08\%) \\
        2 & Urban (low density) & 740326 (5.85) & 307343 (2.43) & 942652 (7.45) \\
        3 & Industrial & 502479 (3.97) & 188150 (1.49) & 649285 (5.13) \\
        4 & Parking & 9198 (0.07) & 2779 (0.02) & 20469 (0.16) \\
        5 & Road & 57634 (0.46) & 8898 (0.07) & 74980 (0.59) \\
        6 & Rapeseed & 79401 (0.63) & 247425 (1.96) & 291462 (2.30) \\
        7 & Wheat \& Barley (winter) & 509236 (4.03) & 773027 (6.11) & 444315 (3.51) \\
        8 & Barley (spring) & 22135 (0.18) & 45397 (0.36) & 81282 (0.64) \\
        9 & Pea & 15298 (0.12) & 103890 (0.82) & 49802 (0.39) \\
        10 & Soy & 6296 (0.05) & 262310 (2.07) & 177084 (1.40) \\
        11 & Sunflower & 298067 (2.36) & 1823222 (14.42) & 106120 (0.84) \\
        12 & Corn & 609941 (4.82) & 305467 (2.42) & 2204111 (17.43) \\
        13 & Corn silage & 827715 (6.54) & 339535 (2.68) & 644489 (5.10) \\
        15 & Beetroot & 207575 (1.64) & 9636 (0.08) & 302543 (2.39) \\
        16 & Potatoes & 26617 (0.21) & 3465 (0.03) & 40855 (0.32) \\
        17 & Grassland & 2277897 (18.01) & 533604 (4.22) & 1119211 (8.85) \\
        18 & Orchards & 527 (<0.01) & 16434 (0.13) & 7337 (0.06) \\
        19 & Vineyards & 2578 (0.02) & 357489 (2.83) & 19145 (0.15) \\
        20 & Deciduous forest & 1088129 (8.60) & 926583 (7.33) & 1972989 (15.60) \\
        21 & Coniferous forest & 4732777 (37.42) & 1091930 (8.63) & 5373252 (42.48) \\
        22 & Lawn & 128711 (1.02) & 682709 (5.4) & 148231 (1.17) \\
        23 & Woodlands & 347466 (2.75) & 368759 (2.92) & 52902 (0.42) \\
        24 & Minerals & 381 (<0.01) & 8483 (0.07) & 2072 (0.02) \\
        27 & Peat & 0 (0) & 0 (0) & 5324 (0.04) \\
        28 & Marshland & 0 (0) & 71985 (0.57) & 7130 (0.06) \\
        30 & Water & 140359 (1.11) & 261436 (2.07) & 375212 (2.97) \\
        \midrule
        \multicolumn{2}{c|}{TOTALS} & 12,647,452 (100) & 8,758,196 (100) & 15,122,125 (100) \\
        \bottomrule
    \end{tabular}
\end{table}

\section{Experiments}
\label{sec:exp}

In the following experiments, a given run was performed by training a model using all of the available source training data and a fixed quantity of target training data.
The target data represents a fixed number of polygons, rather than a fixed amount of data.
A polygon represents a contiguous area with the same land cover---eg. a farm, forest or residential area---meaning that the number of instances in a polygon can be as few as 7 to well over 1,000. 
On average, tile T31TDJ has 336 instances (i.e. time series) per polygon and tile T32ULU has 279 instances per polygon.

Treating the target data in this manner makes the problem more realistic, but also more difficult.
More realistic because in practice reference data is collected per site (polygon), and not per satellite pixel.
More difficult as rather than having an increasing random sample, the data are not distributed across the whole domain and do not represent the accurate class distribution of the area.
For example, if an experiment is performed with 10 polygons of target data this will equate to approximately 3,000 training instances, but it will represent at most 10 classes from the target domain and all the instances within one of these classes will be quite similar.
A sample of this nature is more difficult to learn from than a sample of the same quantity that is randomly selected across the whole target domain. The number of polygons was increased according to the following schedule:
\begin{itemize}
    \item[] \textit{no. of polygons:} $\{1, 2, 4, 8, 16, 32, 64, 128, 256, 512, 1000, 2000\}$
\end{itemize}

Each experiment was repeated five times, and to enable comparison between runs, a linear interpolation of the test accuracies was applied to give the test accuracy for specific quantities of training data (number of pixels).
To enable comparison between methods, the interpolated results from each of the five runs were averaged. All experiments were performed using an implementation in PyTorch 1.3.1~\cite{pytorch}. Our code and the results of the experiments are available at: \repo.

\subsection{Experimental Settings}
\label{sec:settings}
The following section will begin by describing each of the following seven experimental settings that were compared in our experiments:

\begin{itemize}
    \item \methodname;
    \item 4 baseline configurations: Source Only, Target Only, Naive TempCNN, and Finetuned TempCNN;
    \item 2 state-of-the-art semi-supervised DA methods: MiniMax Entropy (MME), and Domain-adversarial Neural Networks (DANN).
\end{itemize}
These configurations are detailed below. 

\subsubsection{\methodname}
\label{subsubsec:ours}
\methodname\ starts with a TempCNN model~(Section~\ref{subsec:tempcnn}) trained on the source domain data.
The weights of this model are used as the initial values for training on the labelled target data with the amount the model is allowed to vary from these values ($\lambda$), based on the quantity of labelled target data.
We note that this highlights a significant benefit of \methodname---that its prerequisites for use are only to have available a pre-trained model and the labelled target data. 
That is, the practitioner does not have to be in possession of the source data to apply our method, as opposed to MME and DANN (presented in Section~\ref{subsubsec:sota}), where all labelled instances (source and target) are pooled and the model is trained on this pooled data, and therefore these methods require all of the labelled source data to be available.
This can be of significant practical benefit as labelled training data from a Sentinel-2 tile is of the order of hundreds of gigabytes and using \methodname\ means that this only has to be stored and used for training on one occasion. 

The value of $\lambda$ is a function of a single hyperparameter: $t_{max}$ (see Section~\ref{subsec:lambda}), which we have set to $10^6$ for all of our experiments.
We believe this to be a reasonable choice as a training set of $10^6$ instances provides enough variation to train an accurate model and thus, it is unlikely that restricting the model's learning by regularizing towards the source parameters will be beneficial to the overall accuracy (we note that a sensitivity analysis is provided in section~\ref{subsubsec:sensitivity}).
Substituting this value into equation~\ref{final_slope}, gives $k = -3.3333$ and the final schedule for our $\lambda$ values as:

\begin{align*}
    \lambda = 10^{10} \cdot\ {n_{t}^{-3.3333}} \numberthis \label{our_lambda} \\
\end{align*}

Once $\lambda$ is calculated the model is trained with SourceRegLoss (Equation~\ref{sourceregloss}) using the Adam optimizer~\cite{kingma2014adam}. We vary the number of epochs used for training with the quantity of training data such that $5,000$ gradient updates have been performed or 1 epoch is completed (see Equation~\ref{epochs}), as due to the large quantity of training data, little learning occurs beyond this point (and this was shown specifically for TempCNN in \citet{Pelletier2019}).

\begin{equation} \label{epochs}
    \text{NoEpochs} =\max\left(1, \frac{\text{NoGradUpdates}\cdot \text{BatchSize}}{\text{TargetTrainQty}}\right)
\end{equation}

\subsubsection{Baseline Configurations}
The following four settings correspond to methods that can map the target domain without using any DA methods, and in doing so, represent various lower bounds for our method to compare to (also see~\citet{Lucas2019multitemp} for a discussion of the performance of baseline CNN configurations).

\paragraph{Source Only}
This is the baseline configuration in which a model is trained on labelled source data only.
This is the simplest setting as it is independent of the amount of labelled target data available, and hence returns only a single value for test accuracy.
This configuration sets the lower bound we would expect for test accuracy when no target data is available and no DA method is applied. For comparative purposes, we have used the TempCNN model, the same categorical cross-entropy loss function, and the same number of training epochs as used for \methodname.

\paragraph{Target Only}
This is a baseline configuration in which the only labelled data used for training the model is from the target domain.
This configuration also acts as a lower bound on the test accuracy for when DA is no longer required, that is, enough target data is available to train an accurate classifier.
As per the \textit{Source only} configuration we have used the TempCNN model, categorical cross-entropy loss function and number of training epochs as we used for \methodname.

\paragraph{Naive TempCNN}
This is a baseline configuration in which a TempCNN model is first trained on the labelled source data and then trained on the labelled target data, without applying a particular DA method.

\paragraph{Finetuned TempCNN}
This is a baseline configuration where the TempCNN is first trained on the labelled source data, at which point the weights of the convolutional layers are frozen, and only the fully-connected layer(s) and the softmax are finetuned by training on the labelled target data.

This technique is common in transfer learning for computer vision problems~\cite{yosinski2014} as the convolutional layers of a CNN learn general features of data while the fully-connected layers learn specifics.
This configuration allows for comparison as to whether the general features of SITS data can first be learnt from a (/any) source domain and then finetuned using the available labelled target data.

\subsubsection{State-of-the-art Methods}
\label{subsubsec:sota}
The two methods presented here represent the current state of the art in semi-supervised DA.
Like \methodname, they are both deep learning-based methods, however unlike our method, each of these require the labelled source data to be available to train the model for the target domain.
These methods concatenate the labelled source and labelled target data and train on this pooled dataset.
While we note that using the labelled data in this manner creates a different problem (an easier one), these methods have been included to illustrate that our method is competitive in accuracy (or indeed outperforming) with the current state of the art, with the additional benefit of only requiring a model pre-trained on the source domain (not all of the source data).
\revision{In each case we attempted to tune the parameters of state-of-the-art method to achieve optimal accuracy for the given method and model architecture.}

\paragraph{MME}
This state-of-the-art method~\cite{saito2019semi} is based on training a CNN model using 2 loss functions.
It learns a prototype of each class from the labelled data and then minimizes the distance between these prototypes and the unlabelled data, in the process learning discriminating features of the data.
It can be implemented on any CNN model, so for comparison purposes we have implemented it using the TempCNN architecture used in our model, so as to control for model choice.
Training occurs in two steps: first, a batch of labelled data (pooled from both source and target domains) is passed forward through the model, a \textit{standard} loss function is calculated and the weights are updated via backpropagation; then, a batch of unlabelled data is passed forward through the model and an entropy loss function is calculated using the output of the convolutional layers of the model.
Once trained, unlabelled target data is tested as per a standard CNN, via a forward pass of the model.
\revision{We trained our MME model using 1 epoch of the labelled training data with a batch size of 32 instances, and the Adam optimizer.
This results in greater than 500,000 gradient updates to the model.}

\paragraph{DANN}
This state-of-the-art method~\cite{ganin2016dann} is based on maximizing accuracy in predicting the class label of an instance while not being able to tell whether it was from the source or target domain.
Learning in this manner discourages the model from learning features that are specific to a domain.
To achieve this it uses a CNN model with two output layers---one for the class and one for the domain.
In our case, each instance has a class label (land cover: 0-29) and a domain label (binary: source/target) and the loss function is the addition of the loss calculated using the class labels and the inverse of the loss calculated using the domain labels.
For unlabelled instances, there will be only the domain label available.
We have used a TempCNN model with two outputs following the convolutional layers, each with a fully connected layer and softmax.
Once trained, unlabelled target data is tested via a forward pass of the model and the class labels are recorded (the domain labels are ignored).
\revision{The DANN model was trained using 1 epoch of the labelled and unlabelled training data with a batch size of 32 instances, and the Adam optimizer.}

\subsection{Results}
The following section presents the results of semi-supervised DA experiments performed on two source-target domain pairs:
\begin{enumerate}
    \item T31TEL~(source)--T31TDJ~(target); and
    \item T31TEL~(source)--T32ULU~(target).
\end{enumerate}

\revision{The results are presented for our method, \methodname, the two state-of-the-art methods, as well as the four baseline configurations.
In each instance, the overall accuracy is presented for an increasing amount of labelled target training data.
The results of our experiments are presented in Tables~\ref{table:overall_acc_TDJ} and \ref{table:overall_acc_ULU}.}

\begin{table}[!htbp]
    \centering
    \caption{\revision{Overall accuracy on target tile T31TDJ for semi-supervised methods and baseline models trained on source tile T31TEL and an increasing quantity of labelled target data.}}
    \label{table:overall_acc_TDJ}
    \begin{tabular}{c|c|c|c|c|c|c}
        \toprule
        \textbf{Target Qty} & \textbf{Naive} & \textbf{Finetuned} & \textbf{Target Only} & \textbf{DANN} & \textbf{MME} & \textbf{Sourcerer} \\
        \midrule
            0 & 0.\textbf{7418} & \textbf{0.7418} & <0.001 & 0.6802 & 0.6671 & \textbf{0.7418} \\
            20 & 0.7415 & 0.6923 & 0.0061 & 0.6856 & 0.6671 & \textbf{0.7417} \\
            50 & 0.7373 & 0.6834 & 0.0150 & 0.6838 & 0.6665 & \textbf{0.7417} \\
            100 & 0.7303 & 0.6830 & 0.0420 & 0.6857 & 0.6689 & \textbf{0.7418} \\
            250 & 0.7076 & 0.6402 & 0.1085 & 0.7042 & 0.6791 & \textbf{0.7425} \\
            500 & 0.6719 & 0.6324 & 0.1267 & 0.6906 & 0.6910 & \textbf{0.7473} \\
            1000 & 0.6461 & 0.6631 & 0.2208 & 0.6951 & 0.7035 & \textbf{0.7537} \\
            5000 & 0.7114 & 0.7373 & 0.4509 & 0.7004 & 0.7202 & \textbf{0.7538} \\
            10000 & 0.7518 & 0.7513 & 0.5387 & 0.6955 & 0.7294 & \textbf{0.7520} \\
            25000 & 0.7623 & 0.7676 & 0.6337 & 0.6941 & 0.7496 & \textbf{0.7680} \\
            50000 & 0.7810 & \textbf{0.7868} & 0.7054 & 0.7104 & 0.7689 & 0.7835 \\
            100000 & 0.8059 & \textbf{0.8063} & 0.7534 & 0.7112 & 0.7845 & 0.8043 \\
            500000 & 0.8434 & 0.8419 & 0.8266 & 0.7491 & 0.8265 & \textbf{0.8455} \\
            1000000 & \textbf{0.8507} & 0.8504 & 0.8419 & 0.7686 & 0.8378 & \textbf{0.8507} \\
        \bottomrule
    \end{tabular}
\end{table}

\begin{table}[!htbp]
    \centering
    \caption{\revision{Overall accuracy on target tile T32ULU for semi-supervised methods and baseline models trained on source tile T31TEL and an increasing quantity of labelled target data.}}
    \label{table:overall_acc_ULU}
    \begin{tabular}{c|c|c|c|c|c|c}
        \toprule
        \textbf{Target Qty} & \textbf{Naive} & \textbf{Finetuned} & \textbf{Target Only} & \textbf{DANN} & \textbf{MME} & \textbf{Sourcerer} \\
        \midrule
            0 & \textbf{0.8747} & \textbf{0.8747} & <0.001 & 0.8411 & 0.8022 & \textbf{0.8747} \\
            20 & 0.8704 & 0.8340 & 0.0020 & 0.8429 & 0.8022 & \textbf{0.8747} \\
            50 & 0.8656 & 0.7854 & 0.0065 & 0.8430 & 0.8040 & \textbf{0.8747} \\
            100 & 0.8582 & 0.7919 & 0.0186 & 0.8396 & 0.8103 & \textbf{0.875} \\
            250 & 0.8341 & 0.7950 & 0.0548 & 0.8380 & 0.8133 & \textbf{0.8751} \\
            500 & 0.8252 & 0.8082 & 0.1046 & 0.8401 & 0.8088 & \textbf{0.8753} \\
            1000 & 0.8128 & 0.8232 & 0.1960 & 0.8431 & 0.8141 & \textbf{0.8753} \\
            5000 & 0.8432 & 0.8320 & 0.5436 & 0.8311 & 0.8367 & \textbf{0.8754} \\
            10000 & 0.8584 & 0.8549 & 0.6646 & 0.8342 & 0.8383 & \textbf{0.8755} \\
            25000 & 0.8609 & 0.8590 & 0.7022 & 0.8368 & 0.8417 & \textbf{0.8776} \\
            50000 & 0.8607 & 0.8609 & 0.7531 & 0.8376 & 0.8491 & \textbf{0.8787} \\
            100000 & 0.8707 & 0.8761 & 0.8108 & 0.8465 & 0.8556 & \textbf{0.8799} \\
            500000 & 0.8977 & \textbf{0.8986} & 0.8733 & 0.8509 & 0.8844 & 0.8942 \\
            1000000 & 0.9033 & 0.9037 & 0.8895 & 0.8655 & 0.8919 & \textbf{0.9054} \\
        \bottomrule
    \end{tabular}
\end{table}

\revision{After analysing these results further, we then discuss the computation time of \methodname\ against the state of the art.
We then present a visual analysis of land cover maps produced in the experiments and then analyse the per-class accuracy of the methods. 
We conclude this section with a sensitivity analysis of the one hyperparameter of our method $t_{max}$.}

\subsubsection{\methodname\ Versus the Baseline Configurations}
\begin{figure}[htb!]
    \centering
    \subfigure[]{\includegraphics[width=.49\linewidth]{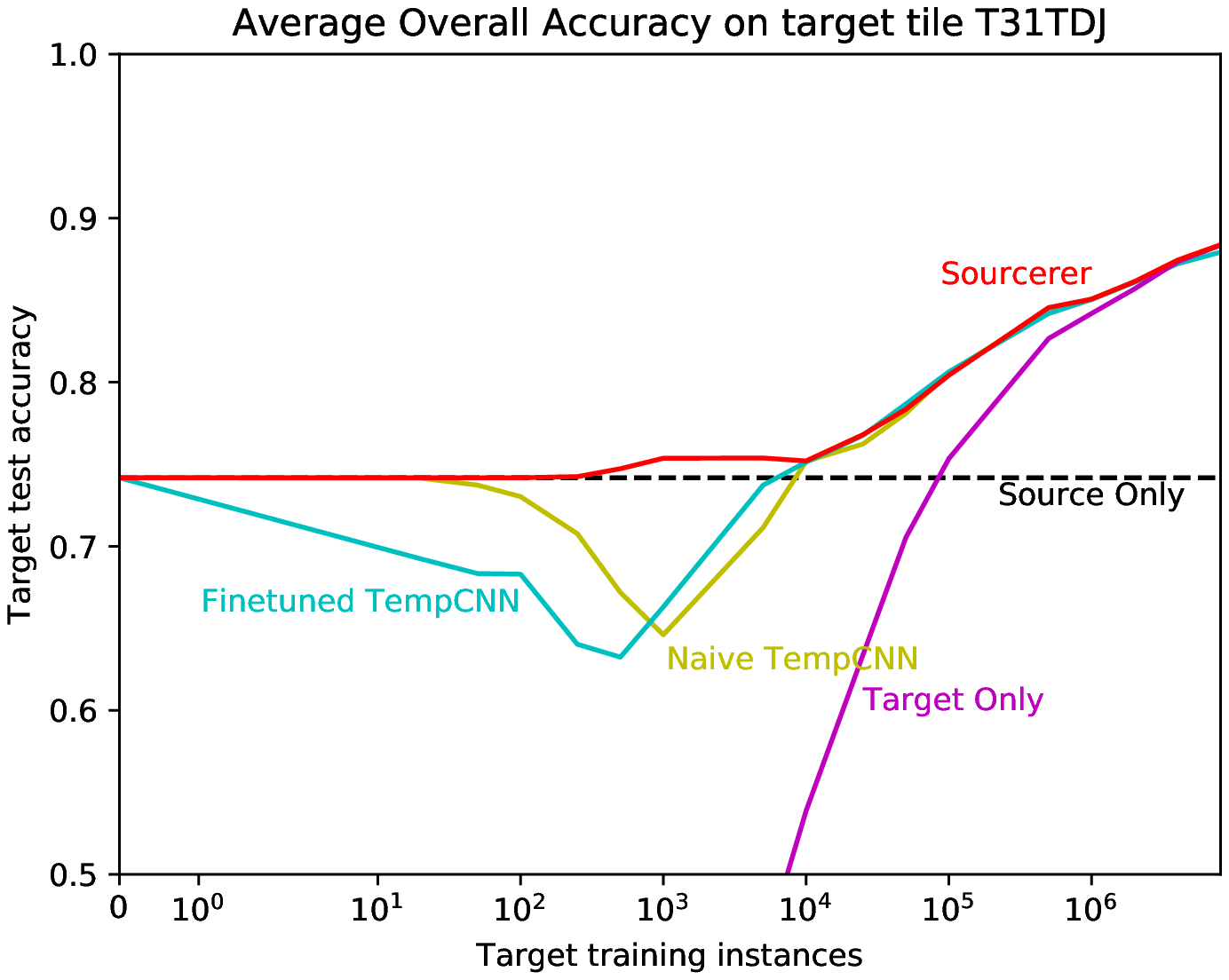}}
    \subfigure[]{\includegraphics[width=.49\linewidth]{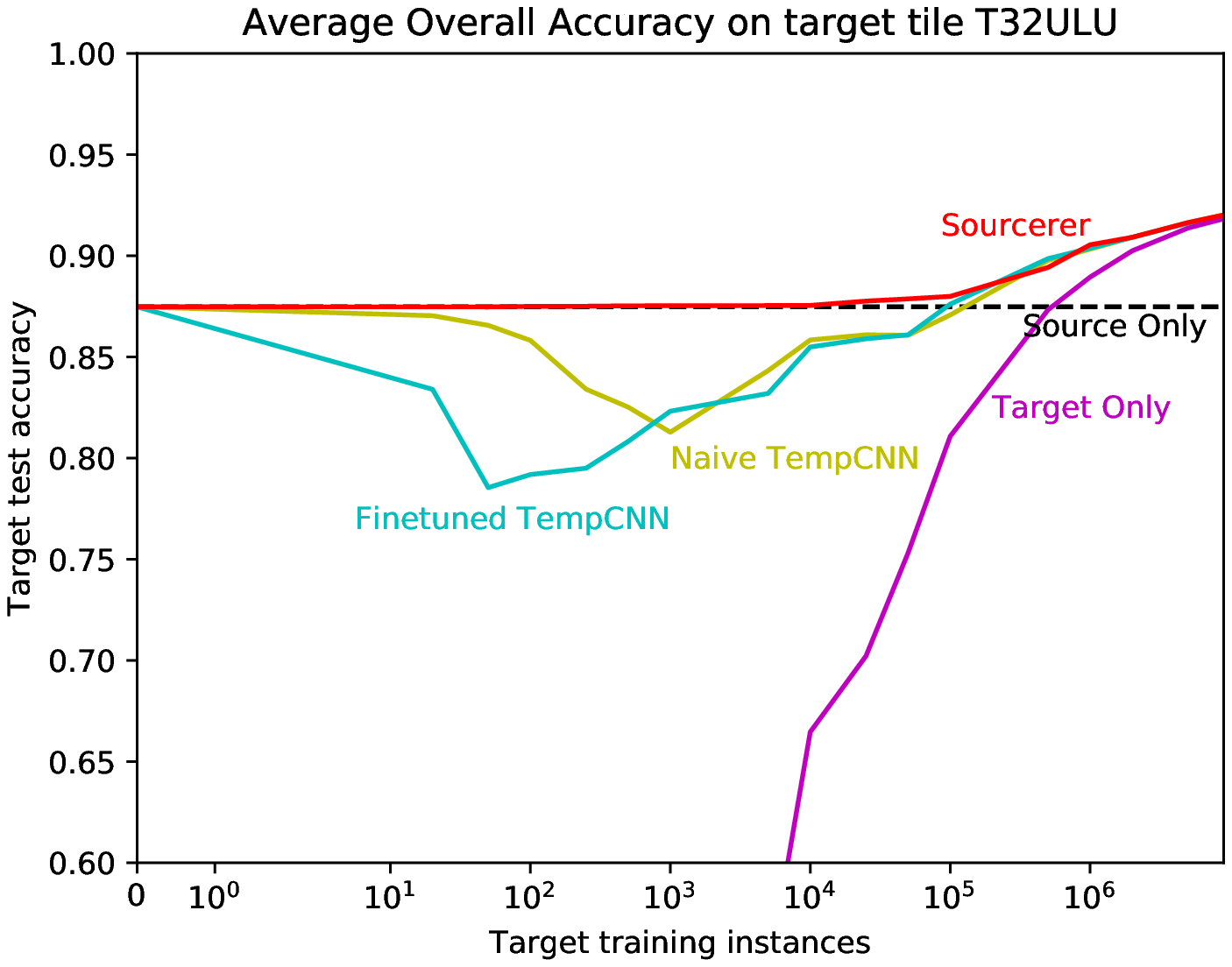}}
    \caption{\label{fig:baseline}Average overall accuracy for \methodname\ against the Baseline configurations, trained on the source domain~(T31TEL) and increasing quantities of labelled target data for domains T31TDJ~(a) and T32ULU~(b).}
\end{figure}

We begin with a comparison of \methodname{} against the baseline configurations.
Figure~\ref{fig:baseline} shows the average overall accuracy for \methodname\ against the baseline configurations--Naive TempCNN, Finetuned TempCNN, Target Only and Source Only--for each target tile.
It shows that for each quantity of target data available, \methodname\ is either equal to or exceeds the performance of all baseline configurations.
This aligns with the intuition of how \methodname\ is designed---for small quantities of target data, the model parameters will be heavily regularized towards those learned on the source data, and hence returns the same accuracy as Source Only; while for large quantities (where DA is not necessary), the model is allowed to learn from the available data and hence returns the same accuracy as Target Only.
The model gradually increases in accuracy between these two extreme situations.

Comparing the performance of \methodname\ on the two tiles, it is evident that the magnitude of its benefit is dependent on the similarity of the source and target domains.
For example, when 25,000 labelled target instances are available \methodname\ outperforms Source Only by 2.5\% on target tile T31TDJ (74.2\% to 76.8\%) where the domains are less similar climatically, compared to 0.3\% (87.5\% to 87.8\%) on tile T32ULU, where the domains are more similar.

An interesting result is also present in the Naive TempCNN and Finetuned TempCNN experiments.
In these setting, it was found that the model initially decreases in accuracy when trained only with labelled target data (see Figure~\ref{fig:baseline}).
On target tile T31TDJ, the Source Only achieves test accuracy of 0.744, while the Naive TempCNN dips to as low as 0.646 when 1,000 labelled target instances are available (approximately 3-4 polygons), before increasing again and growing to be more accurate when moderate-to-large quantities of data are available.
Similarly, tile T32ULU begins at 0.8750 test accuracy and drops to 0.812 before increasing again.
A similar pattern is observed in the Finetuned TempCNN on each target tile.

This dip occurs for two reasons: (1) The available target training data originates from few polygons, and consequently the model overfits the classes present in the target data; and (2) There are some classes present in the target domain that were absent from the source, which significantly shifts the weights of the TempCNN model when they are presented for the first time~\cite{Lucas2019multitemp}.
These results demonstrate that the convolutions of a TempCNN cannot overcome the domain shift alone and that a semi-supervised DA method like \methodname\ is necessary for optimal accuracy.

When considering the Target Only configuration, it takes well over 1M training instances to reach the performance of \methodname\, for each target tile, thus re-emphasizing the case for an accurate semi-supervised DA method.
On target tile T31TDJ, Target Only learns from 100,000 labelled instances before achieving 75\% test accuracy, whereas \methodname\ uses only 1,000.
On tile T32ULU, Target Only requires 500,000 training instances to achieve the starting accuracy of \methodname\ (87.5\%). 

\subsubsection{\methodname\ Versus the State of the Art}
\label{subsec:Sourcerer Versus the State of the Art}
\begin{figure}[htb!]
    \centering
    \subfigure[T31TDJ: x-axis in log scale]{\includegraphics[width=.49\linewidth]{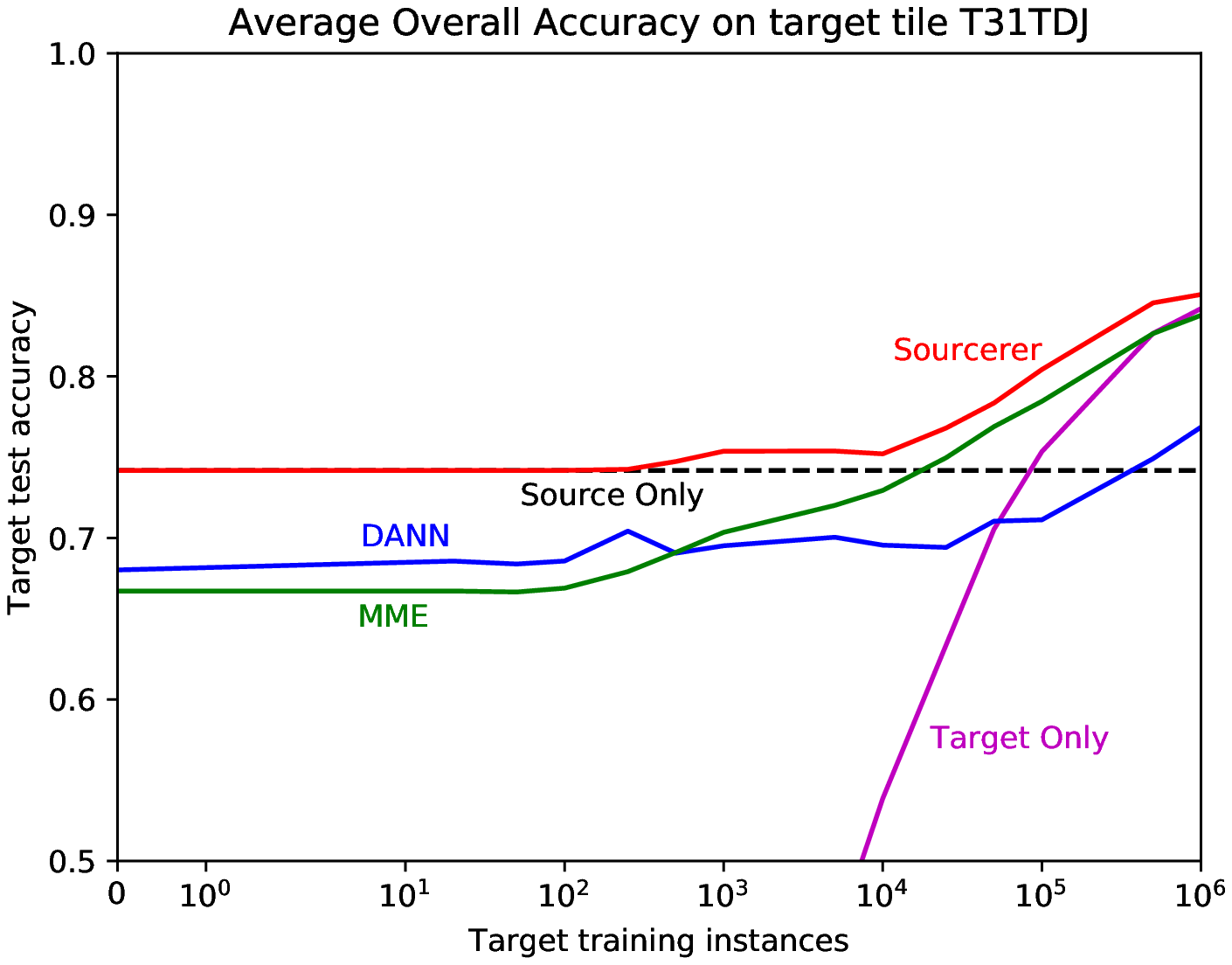}}
    \subfigure[T32ULU: x-axis in log scale]{\includegraphics[width=.49\linewidth]{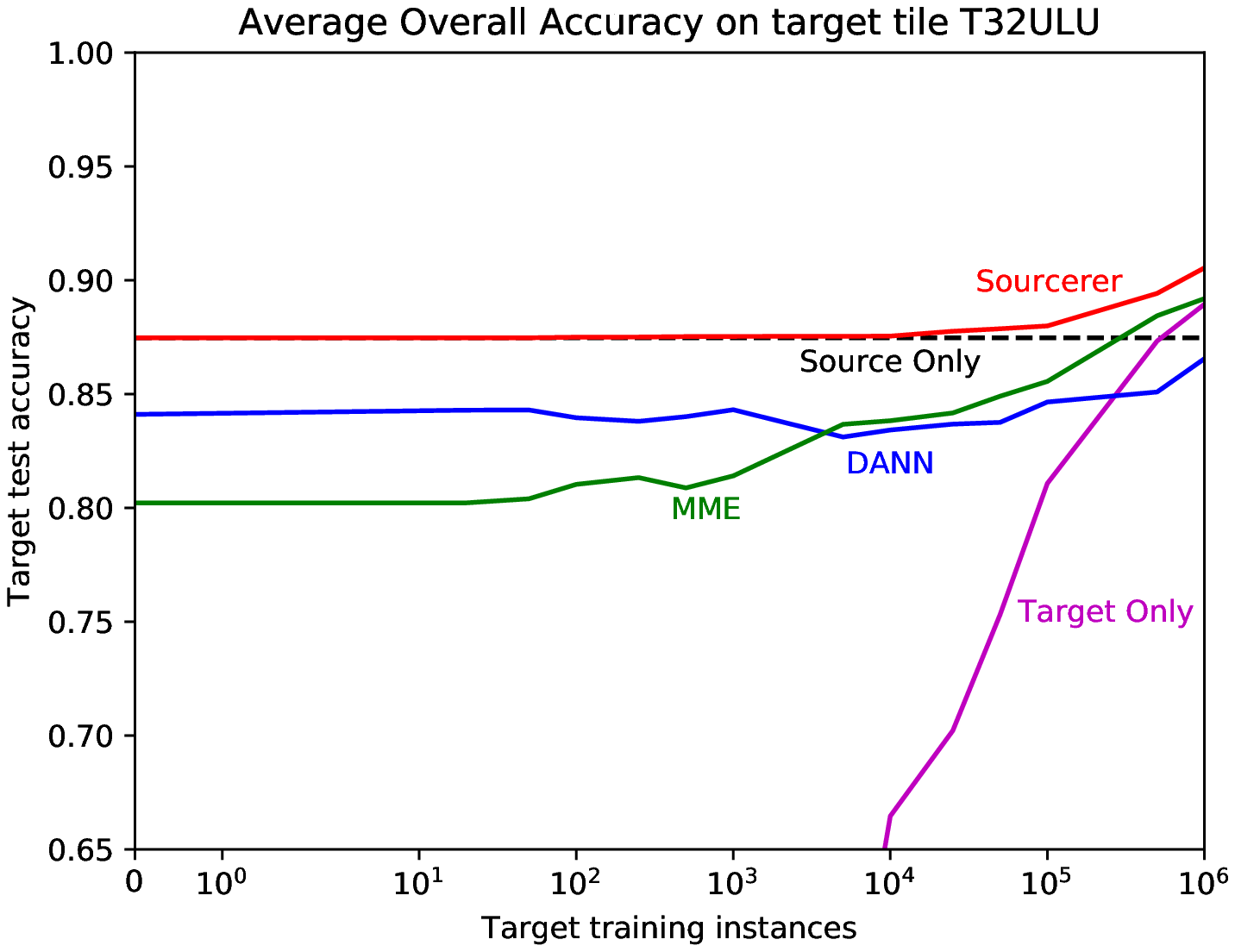}}
    \subfigure[T31TDJ: x-axis in linear scale]{\includegraphics[width=.49\linewidth]{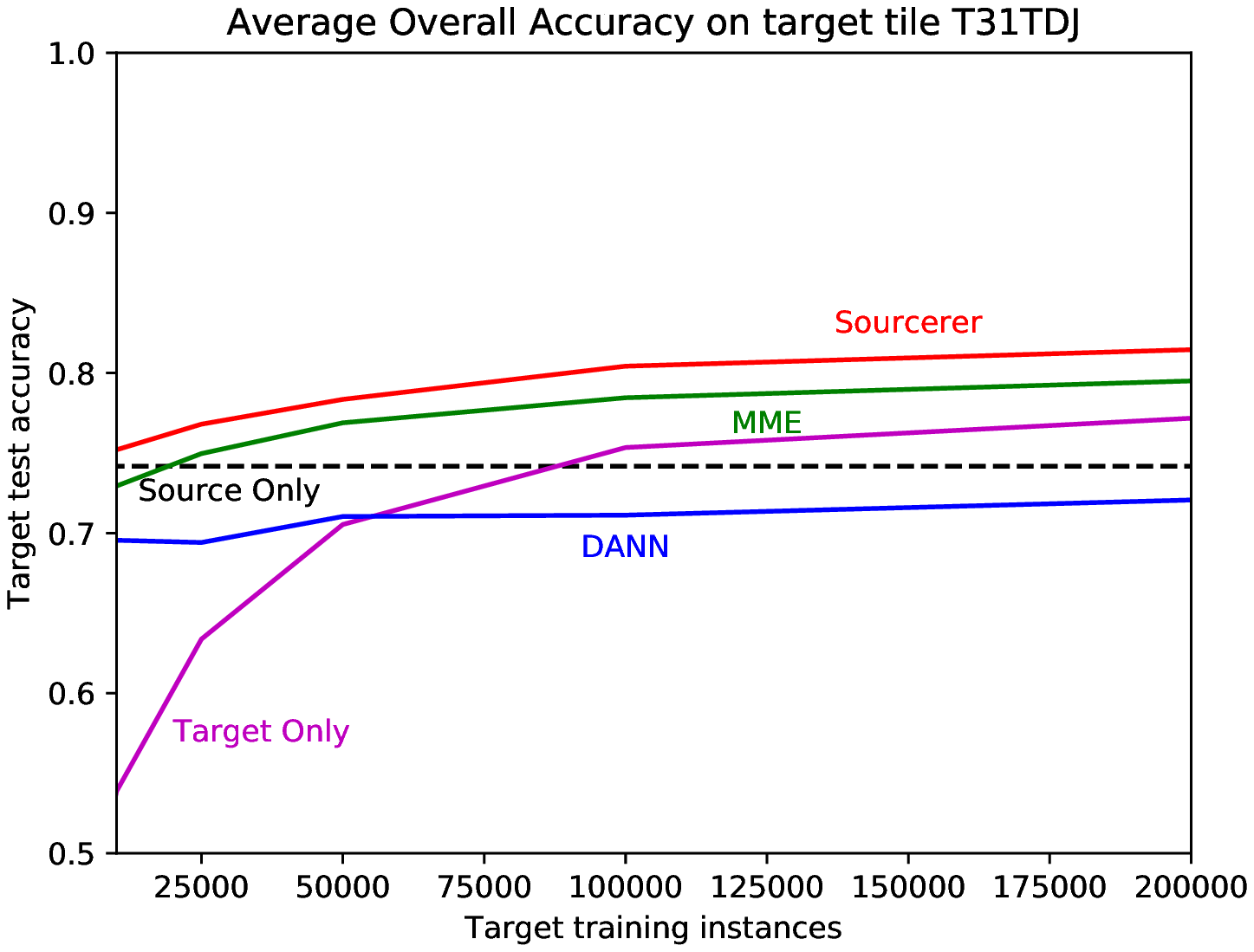}}
    \subfigure[T32ULU: x-axis in linear scale]{\includegraphics[width=.49\linewidth]{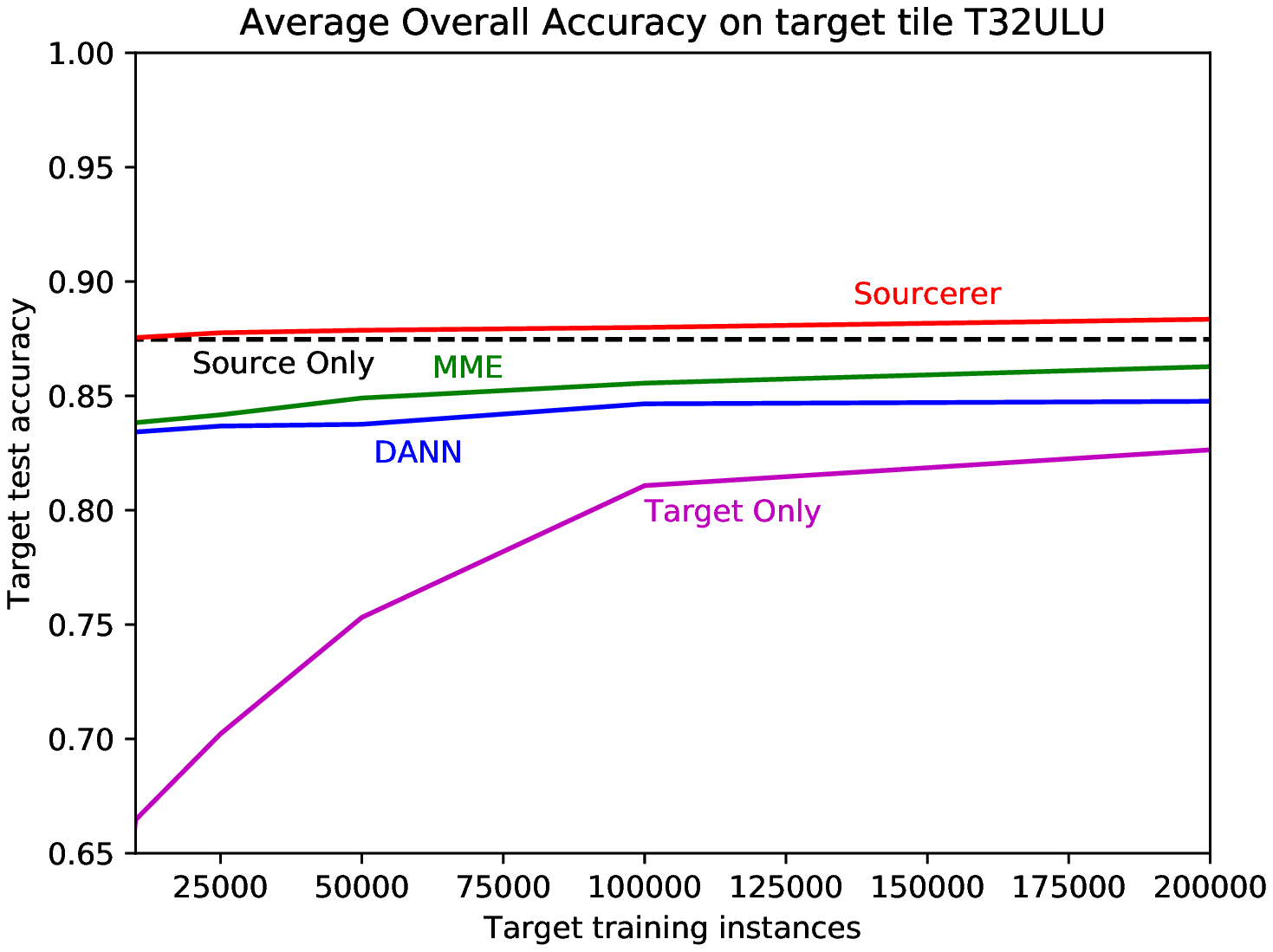}}
    \caption{\label{fig:sota} Average overall accuracy for \methodname\ against the state-of-the-art methods, DANN and MME, and 2 baseline configurations. Models were trained on the source domain~(T31TEL) and increasing quantities of labelled target data. Results are show for target domains T31TDJ in (a)\&(c) and T32ULU in (b)\&(d).}
\end{figure}

We now turn to the most challenging comparison: against MME and DANN.
Figure~\ref{fig:sota} shows the average overall accuracy for \methodname\ against the state-of-the-art methods--DANN and MME--for each target tile.
It is evident from these plots that \methodname, produces a higher test accuracy than either DANN or MME, for any given quantity of labelled target data.
In fact, when considering tile T31TDJ the best possible accuracy achieved by DANN--76.8\% (when training on 1M labelled instances) is achieved by \methodname\ when training on only 25,000 instances.
On tile T32ULU, DANN achieves 86.5\% accuracy using 1M labelled target instances, which is below the initial test accuracy of \methodname\ (87.5\%), that is, without having done any adaptation.

For each experiment, MME starts with the lowest overall accuracy but increases noticeably as more target data become available.
When around 1M target instances are available, it produces a test accuracy within 0.5\% of \methodname, for each target tile.
The improvement indicates that the MME method \textit{is} learning the difference between the domains, however it is not learning quick enough for our application purposes, where greater than 1M instances are unlikely to be available. 

We reiterate that not only is \methodname\ outperforming each of these methods, but it is doing so in a more convenient manner.
Each of MME and DANN use the labelled source data in the training process, whereas once a model is trained on the source data, \methodname\ can use this pre-trained model and the target data to map any target region.

\subsubsection{\revision{Computation Time}}
\revision{The average time for training \methodname\ and the two state-of-the-art configurations is shown in Table~\ref{table:computation_time}. The TempCNN model, which is the starting point for \methodname, completed training on the source domain data in 156.60 minutes.
When training on small quantities of training data, it would take just minutes longer---for example, to train on 1,000 target training instances \methodname\ took only an additional 1.43 minutes on average.
This means that when a source model is available to download, a model can be trained on labelled target data using \methodname\ in minutes.

Interestingly, the models using DANN took the longest time to train when only source data was available, but scarcely increased (and even declined on average) in training time as more labelled target data was used.
This is because each mini-batch used in DANN matches source data with either labelled or unlabelled target data.
That is, each mini-batch of 32 instances is comprised of 16 labelled source instances and 16 target instances (either labelled or unlabelled).
Consequently, as the source data is far larger than the target in quantity, the training time is a function of the amount of source data available.
MME was generally the slowest to train due to the cost of the entropy calculations and performing two gradient updates for each mini-batch. 
The high computation cost is expected as this method is designed to maximise accuracy with very small samples of target data with even class distributions (1-3 instances per class), rather than hundreds of instances from few classes (as in our case).

\begin{table}[!htbp]
    \centering
    \caption{\revision{The average training time (minutes) for Naive TempCNN, \methodname, DANN, and MME, using 12M source instances (Tile T31TEL) and increasing quantities of target data.}}
    \label{table:computation_time}
    \begin{tabular}{c|c|c|c|c}
        \toprule
        \textbf{Target Qty} & \makecell{\textbf{Naive} \\ \textbf{TempCNN}} & \textbf{DANN} & \textbf{MME} & \textbf{\methodname} \\
        \midrule
            0 (source only) & 156.6 & 352.2 & 310.9 & 156.6 \\
            100 & 157.0 & 350.9 & 315.1 & 157.1 \\
            1000 & 157.1 & 369.9 & 383.6 & 158.0 \\
            10000 & 159.2 & 364.9 & 535.9 & 162.0 \\
            100000 & 183.6 & 368.2 & 1301.9 & 185.8 \\
            1000000 & 262.8 & 365.3 & 5737.8 & 268.0 \\
        \bottomrule
    \end{tabular}
\end{table}

The models trained via each method had comparable testing time as the classification process for each is the same---one forward pass of the trained model.
In our experiments, this took approximately 22 minutes on average for target tile T31TDJ (3.4M instances), and 35 minutes on average for target tile T32ULU (5.6M instances).
}

\subsubsection{Visual Analysis of Results} \label{subsubsec:maps}
\begin{figure}
\centering
\includegraphics[width=0.95\linewidth]{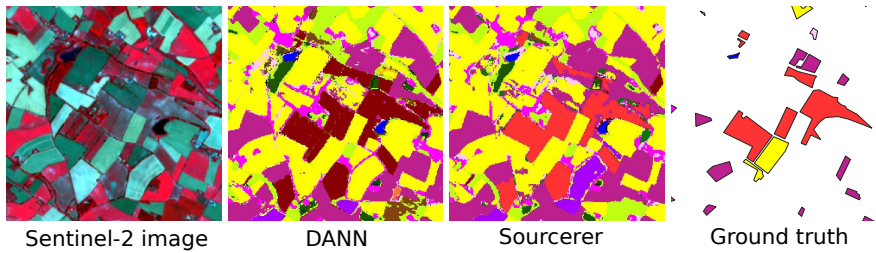}
\caption{A false-color Sentinel-2 image, land cover maps produced using DANN and \methodname\, and the ground truth land cover classes. Maps were created with 64 polygons of labelled target data available (approximately 12,000 instances). Legend provided in Table~\ref{tab:legend}.}
\label{fig:maps_1}
\end{figure}

In this section, we will illustrate what the differences in overall accuracy mean for the resulting land cover maps.
Figure~\ref{fig:maps_1} shows two land cover maps produced by using DANN and \methodname\ and trained on 64 labelled target polygons (approximately 12,000 instances); in comparison with the ground truth polygons from the test data.
When comparing the maps of the two methods, there is disagreement between large areas of agricultural land with the DANN-based model classifying large amounts of corn where \methodname\ classified soy.
As soy and corn are both winter crops, their spectral profiles appear similar and an accurate classifier is required to separate them correctly.
In this case, we can see from the test data that the correct land cover for these polygons are soy as predicted by \methodname.

\begin{figure}[!htb]
\centering
\includegraphics[width=0.95\linewidth]{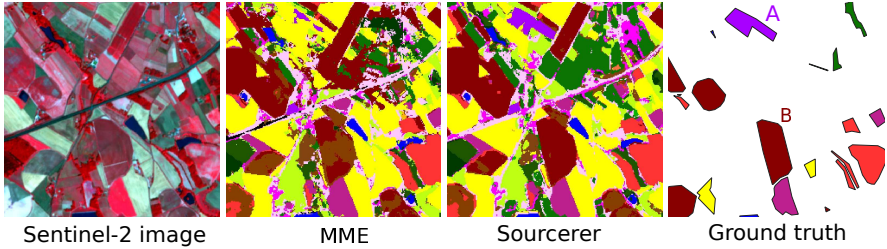}

\caption{A false-color Sentinel-2 image, land cover maps produced using MME and \methodname\, and the ground truth land cover classes. Maps were created with 512 polygons of labelled target data available (approximately 99,000 instances). Legend provided in Table~\ref{tab:legend}.}
\label{fig:maps_2}
\end{figure}

When more data are available the differences between maps produced are more subtle.
Figure~\ref{fig:maps_2} shows land cover maps produced by training on 512 labelled target polygons (approximately 99,000 instances) using MME and \methodname\; as well as a Sentinel-2 false color image and the ground truth.
If we compare the results of each method of classifying the rapeseed crop (crop A in the ground truth subfigure), it can be seen that MME correctly classifies few pixels of this crop while in comparison, \methodname\ accurately classifies almost the whole crop.
When we consider the corn crop (B) located just below the image's center, the MME-based model classifies approximately half of this crop as corn silage, while \methodname\ classifies almost the complete polygon correctly.

\begin{table}
  \centering
  \caption{Legend of land cover classes for Figures \ref{fig:maps_1} and \ref{fig:maps_2} \revision{(only classes present in the data shown)}}
  \label{tab:legend}
  \renewcommand{\arraystretch}{.5}
  \begin{tabular}{clclcl} 
    \hline
    Color & Class & Color & Class & Color & Class \\[4pt] 
    \hline 
        \raisebox{.4\totalheight}{\fcolorbox{black}{Urban high}{}} & Urban (high density) &
        \raisebox{.4\totalheight}{\fcolorbox{black}{Soy}{}} & Soy &
        \raisebox{.4\totalheight}{\fcolorbox{black}{Deciduous}{}} & Deciduous forest \\

        \raisebox{.4\totalheight}{\fcolorbox{black}{Urban low}{}} & Urban (low density) &
        \raisebox{.4\totalheight}{\fcolorbox{black}{Sunflower}{}} & Sunflower &
        \raisebox{.4\totalheight}{\fcolorbox{black}{Coniferous}{}} & Coniferous forest \\
        
        \raisebox{.4\totalheight}{\fcolorbox{black}{Industrial}{}} & Industrial &
        \raisebox{.4\totalheight}{\fcolorbox{black}{Corn}{}} & Corn &
        \raisebox{.4\totalheight}{\fcolorbox{black}{Lawn}{}} & Lawn \\
        
        \raisebox{.4\totalheight}{\fcolorbox{black}{Parking}{}} & Parking &
        \raisebox{.4\totalheight}{\fcolorbox{black}{Corn silage}{}} & Corn silage &
        \raisebox{.4\totalheight}{\fcolorbox{black}{Woodlands}{}} & Woodlands \\
        
        \raisebox{.4\totalheight}{\fcolorbox{black}{Road}{}} & Road &
        \raisebox{.4\totalheight}{\fcolorbox{black}{Beetroot}{}} & Beetroot &
        \raisebox{.4\totalheight}{\fcolorbox{black}{Minerals}{}} & Minerals \\
        
        \raisebox{.4\totalheight}{\fcolorbox{black}{Rapeseed}{}} & Rapeseed &
        \raisebox{.4\totalheight}{\fcolorbox{black}{Potatoes}{}} & Potatoes &
        \raisebox{.4\totalheight}{\fcolorbox{black}{Peat}{}} & Peat \\
        
        \raisebox{.4\totalheight}{\fcolorbox{black}{Wheat Barley}{}} & Wheat \& Barley &
        \raisebox{.4\totalheight}{\fcolorbox{black}{Grassland}{}} & Grassland &
        \raisebox{.4\totalheight}{\fcolorbox{black}{Marshland}{}} & Marshland \\
        
        \raisebox{.4\totalheight}{\fcolorbox{black}{Barley}{}} & Barley (spring) &
        \raisebox{.4\totalheight}{\fcolorbox{black}{Orchards}{}} & Orchards &
        \raisebox{.4\totalheight}{\fcolorbox{black}{Water}{}} & Water \\
        
        \raisebox{.4\totalheight}{\fcolorbox{black}{Peas}{}} & Peas &
        \raisebox{.4\totalheight}{\fcolorbox{black}{Vineyards}{}} & Vineyards  \\
  
  \end{tabular}
\end{table}

\subsubsection{\revision{F1-score Accuracy}}
\revision{The above comparisons all centre around overall classification accuracy to compare methods, and while this is the most commonly used accuracy measure, it is far from the only one.
In this section, we present another accuracy measure, the F1-score, to analyse the results of our experiments.
The average F1-score is calculated by:

\begin{align*}
    F1 = \frac{1}{|classes|} \mathlarger{\mathlarger{\sum}}_{x \in classes}2\frac{precision_x \cdot recall_x}{precision_x + recall_x}
\end{align*}

The F1-scores for \methodname, the two state of the art, and the three baseline configurations are shown in Tables~\ref{table:F1_TDJ} and~\ref{table:F1_ULU}.}

\begin{table}[!htbp]
    \centering
    \caption{\revision{F1-score on target tile T31TDJ for semi-supervised methods and baseline models trained on source tile T31TEL and an increasing quantity of labelled target data.}}
    \label{table:F1_TDJ}
    \begin{tabular}{c|c|c|c|c|c|c}
        \toprule
        \textbf{Target Qty} & \textbf{Naive} & \textbf{Finetuned} & \textbf{Target Only} & \textbf{DANN} & \textbf{MME} & \textbf{Sourcerer} \\
        \midrule
            0 & \textbf{0.4183} & \textbf{0.4183} & <0.001 & 0.3690 & 0.3364 & \textbf{0.4183}\\
            20 & \textbf{0.4279} & 0.3736 & 0.0006 & 0.3700 & 0.3364 & 0.4273\\
            50 & 0.4246 & 0.3661 & 0.0016 & 0.3678 & 0.3391 & \textbf{0.4282}\\
            100 & 0.4163 & 0.3647 & 0.0034 & 0.3668 & 0.3462 & \textbf{0.4184}\\
            250 & 0.3926 & 0.3228 & 0.0092 & 0.3698 & 0.3499 & \textbf{0.4193}\\
            500 & 0.3845 & 0.3240 & 0.0231 & 0.3738 & 0.3704 & \textbf{0.4283}\\
            1000 & 0.3650 & 0.3445 & 0.0614 & 0.3754 & 0.3805 & \textbf{0.4336}\\
            5000 & 0.4064 & 0.4143 & 0.2014 & 0.3651 & 0.4094 & \textbf{0.4418}\\
            10000 & 0.4238 & 0.4191 & 0.2357 & 0.3660 & 0.4166 & \textbf{0.4307}\\
            25000 & 0.4607 & 0.4652 & 0.3199 & 0.3684 & 0.4414 & \textbf{0.4695}\\
            50000 & 0.4707 & \textbf{0.4757} & 0.3912 & 0.3811 & 0.4756 & 0.4756\\
            100000 & 0.4920 & 0.4922 & 0.4360 & 0.3803 & 0.4841 & \textbf{0.4936}\\
            500000 & 0.5559 & 0.5468 & 0.5345 & 0.4324 & 0.5495 & \textbf{0.5588}\\
            1000000 & 0.5707 & 0.5686 & 0.5612 & 0.4564 & 0.5471 & \textbf{0.5760}\\
        \bottomrule
    \end{tabular}
\end{table}

\begin{table}[!htbp]
    \centering
    \caption{\revision{F1-score on target tile T32ULU for semi-supervised methods and baseline models trained on source tile T31TEL and an increasing quantity of labelled target data.}}
    \label{table:F1_ULU}
    \begin{tabular}{c|c|c|c|c|c|c}
        \toprule
        \textbf{Target Qty} & \textbf{Naive} & \textbf{Finetuned} & \textbf{Target Only} & \textbf{DANN} & \textbf{MME} & \textbf{Sourcerer} \\
        \midrule
            0 & \textbf{0.5897} & \textbf{0.5897} & <0.001 & 0.5064 & 0.5151 & \textbf{0.5897}\\
            20 & 0.5824 & 0.5507 & 0.0006 & 0.5012 & 0.5151 & \textbf{0.5897}\\
            50 & 0.5768 & 0.4650 & 0.0013 & 0.5003 & 0.5186 & \textbf{0.5897}\\
            100 & 0.5708 & 0.4783 & 0.0028 & 0.4949 & 0.5164 & \textbf{0.5896}\\
            250 & 0.5470 & 0.5041 & 0.0084 & 0.4815 & 0.5087 & \textbf{0.5891}\\
            500 & 0.5228 & 0.4953 & 0.0251 & 0.4955 & 0.4958 & \textbf{0.5884}\\
            1000 & 0.4838 & 0.5098 & 0.0555 & 0.5010 & 0.5129 & \textbf{0.5867}\\
            5000 & 0.5298 & 0.5144 & 0.2019 & 0.4961 & 0.5276 & \textbf{0.5681}\\
            10000 & 0.5359 & 0.5423 & 0.2778 & 0.4929 & 0.5462 & \textbf{0.5691}\\
            25000 & 0.5417 & 0.5450 & 0.3396 & 0.5040 & 0.5362 & \textbf{0.5503}\\
            50000 & 0.5574 & 0.5612 & 0.3963 & 0.5005 & 0.5514 & \textbf{0.5652}\\
            100000 & 0.5764 & \textbf{0.5829} & 0.4373 & 0.4998 & 0.5539 & 0.5821\\
            500000 & 0.6449 & 0.6493 & 0.5877 & 0.5114 & 0.6301 & \textbf{0.6494}\\
            1000000 & 0.6605 & 0.6614 & 0.6190 & 0.5374 & 0.6435 & \textbf{0.6632}\\
        \bottomrule
    \end{tabular}
\end{table}

\revision{In general, \methodname\ outperforms all of the baselines methods and the state of the art in terms of F1-score accuracy.
However the F1-scores are significantly lower than the overall accuracy values for a given experiment and method---eg. when training on 1,000 labelled T31TDJ instances, \methodname\ achieves an overall accuracy of 0.7520 but only an F1-score of 0.4307.
This indicates that the methods are poorly predicting some classes in the target domain.

A review of the per-class F1-scores indicates that \methodname\ performs less accurately on the classes where the distributions differ between the source and target domains.
For example, Coniferous Forest covers 32.42\% of the source domain but only 8.63\% of target domain T31TDJ (see Table~\ref{tab:class_dist}).
Table~\ref{table:conif} shows the Coniferous Forest F1-score for \methodname\ and the state of the art for one shuffle of the labelled training data from target tile T31TDJ.

We note that the remaining F1-scores for each individual class are available at: \repo{}.
}

\begin{table}[!htbp]
    \centering
    \caption{\revision{F1-score for Coniferous Forest class for a model trained using \methodname, DANN, and MME. These results represent one shuffle of the labelled training data from target tile T31TDJ, while the source model has been trained on tile T31TEL.}}
    \label{table:conif}
    \begin{tabular}{c|c|c"c|c|c}
        \toprule
        \multicolumn{3}{c"}{\textbf{Target Domain}} & \multicolumn{3}{c}{\textbf{F1-score (Conif. Forest)}} \\
        \midrule
        \makecell{Labelled \\ Polygons} & \makecell{Labelled \\ Quantity} & \makecell{Conif. Forest \\ Quantity} & MME & DANN & \methodname \\
        \midrule
            1 & 297 & 297 & \textbf{0.0390} & 0.0001 & 0.0006 \\
            2 & 394 & 297 & \textbf{0.0252} & <0.0001 & 0.0005 \\
            4 & 644 & 297 & \textbf{0.0230} & 0.0001 & 0.0002 \\
            8 & 1198 & 297 & \textbf{0.0154} & <0.0001 & 0.0003 \\
            16 & 1786 & 297 & \textbf{0.1230} & <0.0001 & 0.0198 \\
            32 & 4836 & 297 & \textbf{0.1178} & <0.0001 & 0.0427 \\
            64 & 29195 & 10884 & \textbf{0.0450} & 0.0001 & 0.0087 \\
            128 & 42504 & 10884 & \textbf{0.0451} & 0.0002 & 0.0012 \\
            256 & 73688 & 12498 & \textbf{0.0992} & 0.0001 & 0.0714 \\
            512 & 172746 & 47467 & \textbf{0.0875} & 0.0005 & 0.0415 \\
            1000 & 283090 & 53251 & 0.2762 & 0.0001 & \textbf{0.3482} \\
            2000 & 546778 & 82507 & 0.3020 & 0.0002 & \textbf{0.3623} \\
            4000 & 1094517 & 152598 & 0.3097 & 0.0002 & \textbf{0.4102} \\
        \bottomrule
    \end{tabular}
\end{table}

\revision{These results highlight a limitation of both our model and the state of the art, as all methods perform poorly on this class.
This issue is not specific to our problem however, as learning from unbalanced data is a large area of research in machine learning (see for example: \citet{krawczyk2016}).
One note that we can take away from these results is that the choice of source domain is important for optimal results when using \methodname.}

\subsubsection{Sensitivity Analysis of $t_{max}$}
\label{subsubsec:sensitivity}
\begin{figure}[htb!]
\centering
\subfigure[]{\includegraphics[width=.49\linewidth]{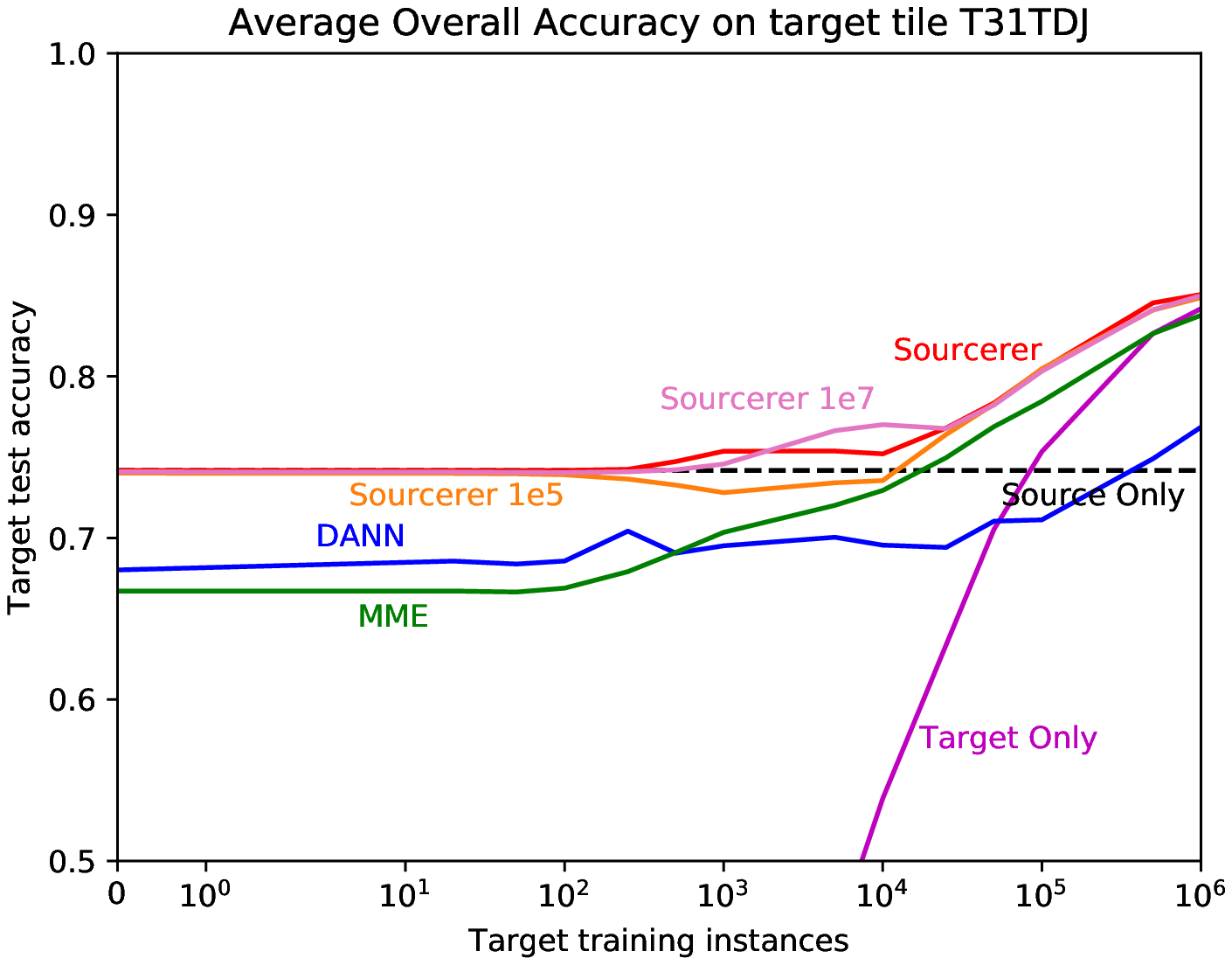}}
\subfigure[]{\includegraphics[width=.49\linewidth]{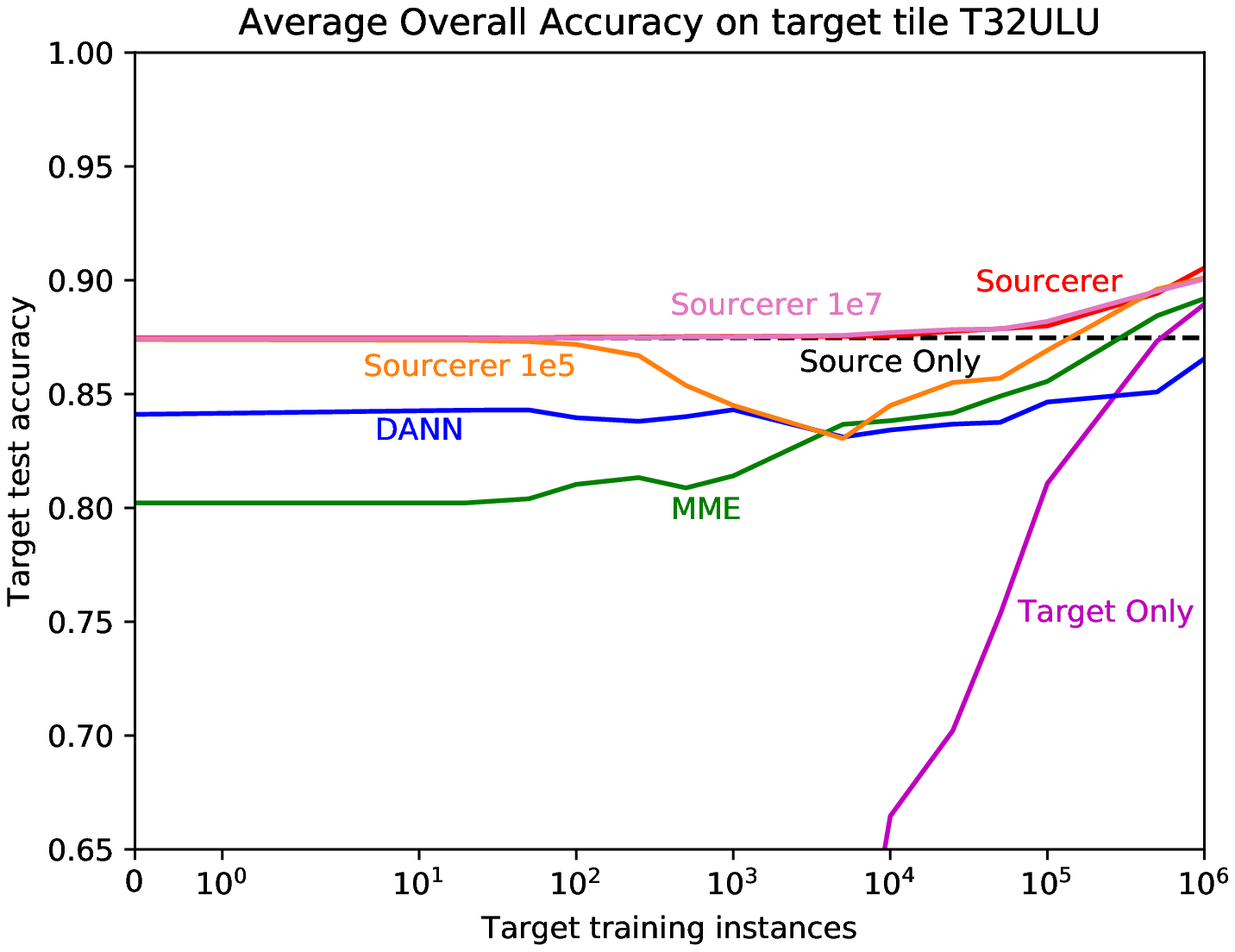}}
\caption{\label{fig:sensitivity}Average overall accuracy for \methodname\ with different values for the hyperparameter $t_{max}$, trained on the source domain~(T31TEL) and increasing quantities of labelled target data for domains T31TDJ~(a) and T32ULU~(b).}
\end{figure}

\methodname\ has only one user-defined hyperparameter, $t_{max}$, which represents the quantity of labelled target data at which the regularization applied to the model approaches zero (as discussed in Section~\ref{subsec:lambda})---~it represents the quantity of target data at which we would no longer require source data and DA to learn an accurate model.
We have performed experiments on each target tile with three different values of $t_{max}$---$10^5$, $10^6$ (the default value), and $10^7$.
Figure~\ref{fig:sensitivity} shows the average overall accuracy for the three models for each target tile.

On tile T31TDJ, each of the three models of \methodname\ outperform the state of the art for all quantities of labelled target data.
This is also the case for tile T32ULU for models with $t_{max}$ of $10^6$ and $10^7$. The model with $t_{max}$ of $10^5$ does dip below the performance of MME when around 8,000 target instances are used.
This \textit{dip} in performance is the same as that displayed by the Naive TempCNN in Section~\ref{subsec:Sourcerer Versus the State of the Art}, and indicates that a value of $10^5$ for the $t_{max}$ does not regularize the model sufficiently.

The results for the other two models show that the choice of $t_{max}$ being either $10^6$ or $10^7$ will produce similar performance, and thus any attempt to optimize this value further is not likely to be necessary.

\section{Conclusion and Future Work}
\label{sec:con}
In this paper we presented \methodname, a Bayesian-inspired, deep learning-based, semi-supervised DA technique for producing land cover maps from SITS data. The technique takes a CNN trained on a source domain and treats this as a prior distribution for the weights of the model, with the degree to which the model is modified to fit the target domain limited by the quantity of labelled target data available.

Our experiments using Sentinel-2 time series images showed that \methodname\ outperforms all other methods for any quantity of labelled target data available on two different source-target domain pairings.
On the more difficult target domain, the starting accuracy (when no labelled target data are available) of \methodname\ is 74.2\%, and this is greater than the next-best state-of-the-art method when trained on 20,000 labelled target instances.

\methodname's high accuracy is also complemented by its straight-forward manner of application as it only requires a model pre-trained of the source domain, rather than all of the source data.
\revision{In this case, a model can be trained on 10,000 labelled target instances using \methodname\ in under six minutes.
This offers great promise to efficiently map resource-poor areas as the practitioner only has to download a model, not millions of instances of source domain data, and spend only a very short time training on the target domain data.}



\revision{In the future, we would like to see how \methodname\ performs in other DA contexts, in particular temporal DA, used to update maps with recently-acquired data.
We would also like to experiment with using \methodname\ across domains with different resolutions or modes of acquisition.

}

\section*{Supplementary material}
To aid replication, the code for our method and the raw results of all experiments are available at \repo{}.

\section*{Acknowledgements}
The authors would like to thank our colleagues from the CESBIO laboratory (in particular Jordi Inglada, Olivier Hagolle, and Arthur Vincent) for providing us with the corrected Sentinel-2 data and associated labels. \revision{We are grateful to the editor and anonymous reviewers whose suggestions and comments have greatly improved the paper.} \\
This research was supported by the Australian Research Council under grant DE170100037. \\

\end{document}